\documentclass[letterpaper]{article} 
\usepackage{aaai2026}  
\usepackage{times}  
\usepackage{helvet}  
\usepackage{courier}  
\usepackage[hyphens]{url}  
\usepackage{graphicx} 
\usepackage{natbib}  
\usepackage{caption} 
\usepackage{varwidth}
\usepackage{graphicx}
\usepackage{subcaption}
\usepackage{amsmath,amssymb,bm}
\usepackage{algorithmicx}
\usepackage{algorithm}

\usepackage{algpseudocode}
\algrenewcommand\alglinenumber[1]{\scriptsize #1} 
\usepackage{tocloft}
\usepackage{microtype}
\usepackage{booktabs} 
\usepackage{amsmath,amssymb,amsfonts}
\usepackage{textcomp}
\usepackage{xcolor}
\usepackage{multirow} 
\usepackage{siunitx}
\usepackage{newtxtext}
\usepackage{pifont}
\usepackage{subcaption}

\allowdisplaybreaks

\usepackage{amsthm}
\usepackage{bm}
\usepackage{array}
\definecolor{lightblue}{RGB}{51,102,255}
\newcommand{\cmark}{\ding{51}}
\newcommand{\xmark}{\ding{55}}

\usepackage{mathtools}
\usepackage{newfloat}
\usepackage{listings}
\DeclareCaptionStyle{ruled}{labelfont=normalfont,labelsep=colon,strut=off} 
\floatstyle{ruled}
\newfloat{listing}{tb}{lst}{}
\floatname{listing}{Listing}
\title{SimDiff: Simpler Yet Better Diffusion Model for Time Series Point Forecasting}
\author {
    Hang Ding$^{1,}$\thanks{ Equal contribution \\ \quad $^{\dagger}$ Corresponding author}
    Xue Wang$^{2,*}$,
    Tian Zhou$^{3,4,*}$,
    Tao Yao$^{1,\dagger}$
}
\affiliations {
    \textsuperscript{\rm 1}Shanghai Jiao Tong University\\
    \textsuperscript{\rm 2}Alibaba Group US\\
    \textsuperscript{\rm 3}DAMO Academy, Alibaba Group\\
    \textsuperscript{\rm 4}Hupan Lab\\
    dearsloth@sjtu.edu.cn, taoyao@sjtu.edu.cn
}

\usepackage{bibentry}

\begin{document}

\maketitle

\begin{abstract}
Diffusion models have recently shown promise in time series forecasting, particularly for probabilistic predictions. However, they often fail to achieve state-of-the-art point estimation performance compared to regression-based methods. This limitation stems from difficulties in providing sufficient contextual bias to track distribution shifts and in balancing output diversity with the stability and precision required for point forecasts. Existing diffusion-based approaches mainly focus on full-distribution modeling under probabilistic frameworks, often with likelihood maximization objectives, while paying little attention to dedicated strategies for high-accuracy point estimation. Moreover, other existing point prediction diffusion methods frequently rely on pre-trained or jointly trained mature models for contextual bias, sacrificing the generative flexibility of diffusion models.

To address these challenges, we propose SimDiff, a single-stage, end-to-end framework. SimDiff employs a single unified Transformer network carefully tailored to serve as both denoiser and predictor, eliminating the need for external pre-trained or jointly trained regressors. It achieves state-of-the-art point estimation performance by leveraging intrinsic output diversity and improving mean squared error accuracy through multiple inference ensembling. Key innovations, including normalization independence and the median-of-means estimator, further enhance adaptability and stability. Extensive experiments demonstrate that SimDiff significantly outperforms existing methods in time series point forecasting. 
\end{abstract}

\begin{links}
    \link{Code}{https://github.com/Dear-Sloth/SimDiff/tree/main}
\end{links}

\newcommand{\tcr}{\textcolor{red}}
\newcommand{\tcb}{\textcolor{blue}}
\newcommand{\modelname}{MotifCAR}

\section{Introduction}

Time series forecasting plays a crucial role across various real-world domains, including economics~\cite{Friedman1962}, retail sales prediction ~\cite{bose2017probabilistic,courty1999timing,qiu2024tfb}, and energy management~\cite{gao2020robusttad,qiu2025tab}. The fundamental process of time series forecasting involves generating future sequences based on historical observations, which is intuitively well-suited for the application of diffusion models.
\begin{figure}[ht]
    \centering
    \includegraphics[width=0.9\linewidth]{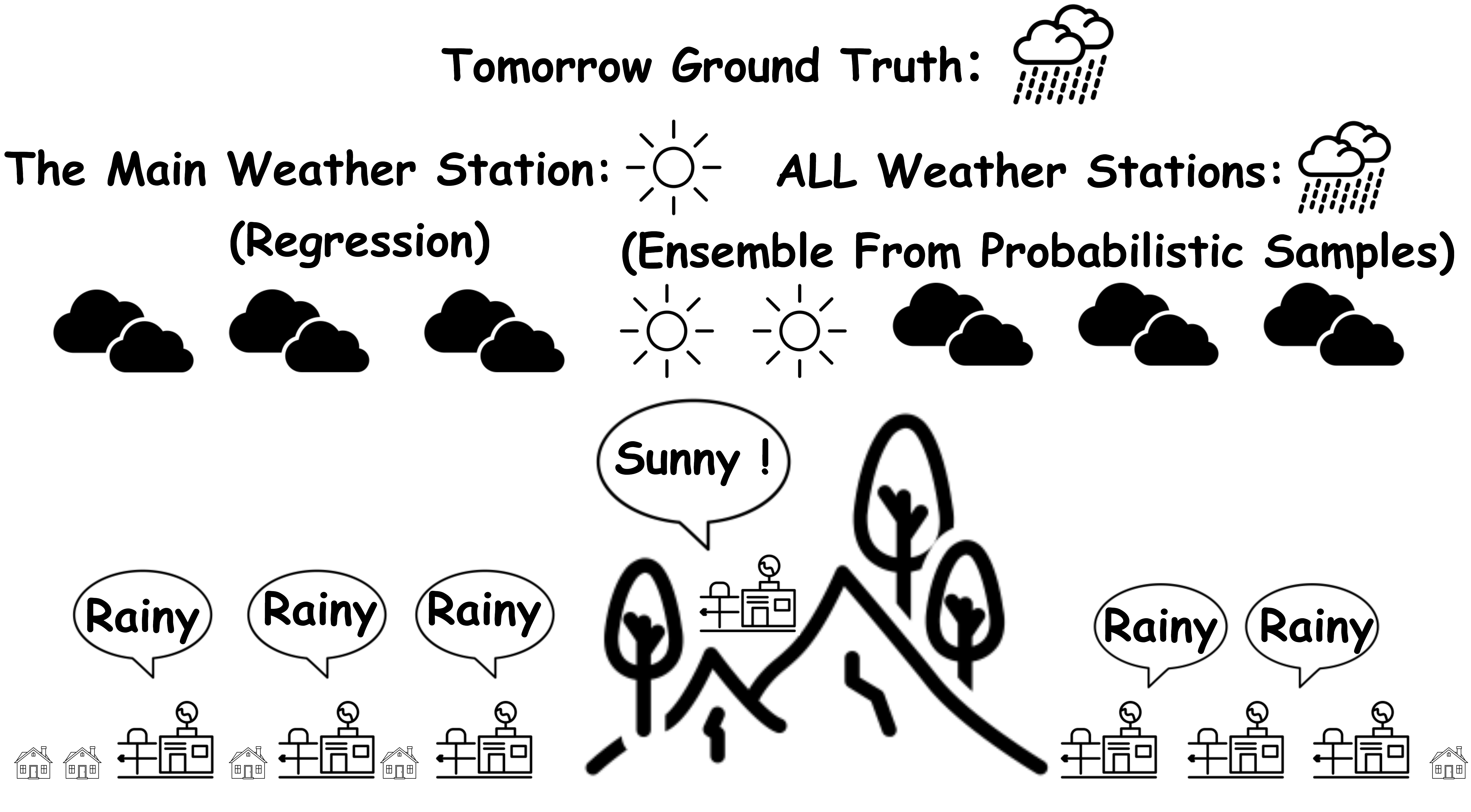}
    \caption{Trustworthy Forecasting by Ensembling Diverse Probability Samples}
    \label{fig:ens}
\end{figure}

\begin{figure*}[ht]
    \centering
    \begin{subfigure}{0.3\textwidth}
        \centering
        \includegraphics[width=\linewidth]{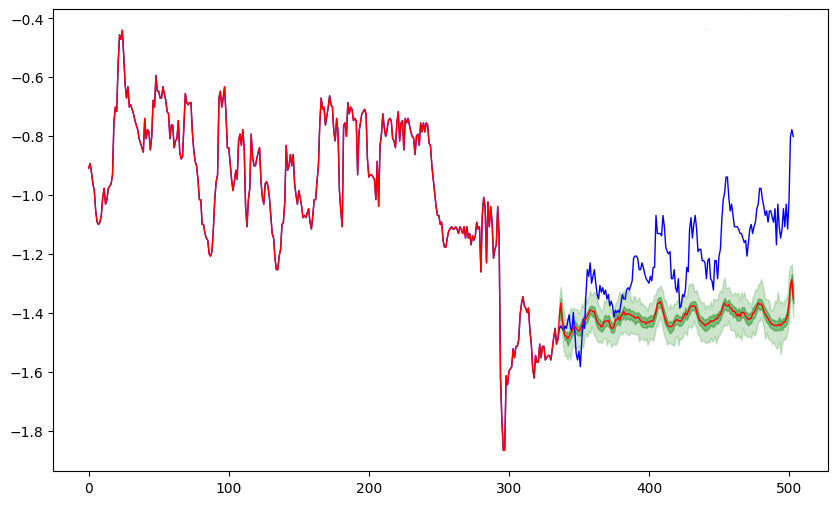}
        \caption{CSDI}
        \label{fig:csdi1}
    \end{subfigure}
    \hfill
    \begin{subfigure}{0.3\textwidth}
        \centering
        \includegraphics[width=\linewidth]{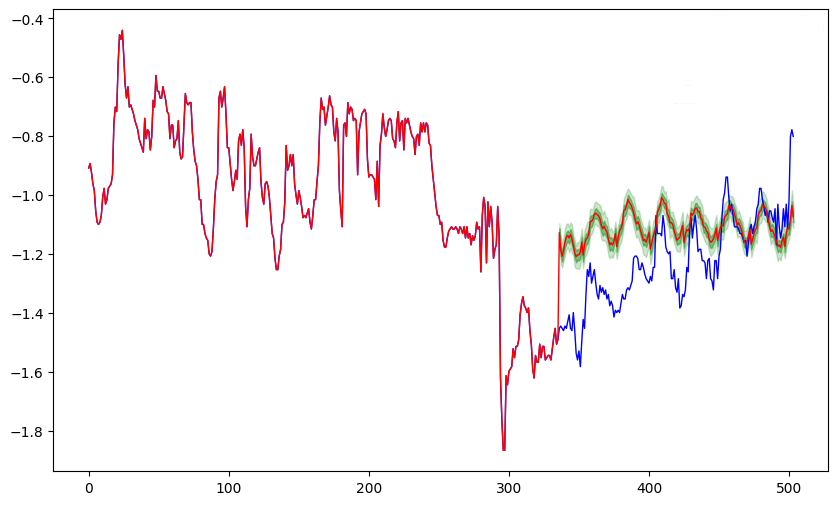}
        \caption{TimeDiff}
        \label{fig:timediff1}
    \end{subfigure}
    \hfill
    \begin{subfigure}{0.3\textwidth}
        \centering
        \includegraphics[width=\linewidth]{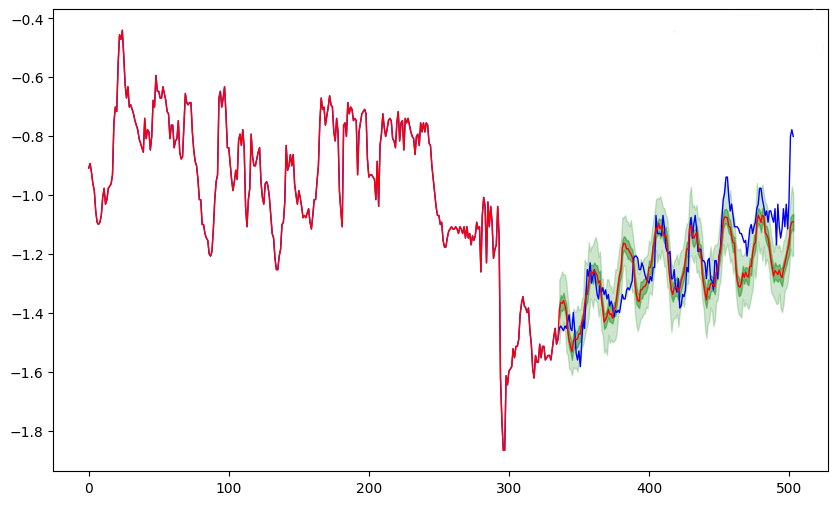}
        \caption{SimDiff}
        \label{fig:simdiff1}
    \end{subfigure}
    
    \caption{Visualizations on \textit{ETTh1} by (a) \textit{CSDI}, (b) \textit{TimeDiff}, and (c) \textit{SimDiff}. CSDI only shows 90\% interval due to the existence of extreme samples.}
    \label{fig:comparison}
\end{figure*}

Two inseparable questions continue to hinder diffusion-based forecasting: \textit{(i) how to inject sufficient contextual bias from past observations to obtain a stable and faithful predictive distribution? and (ii) how to reconcile the inherent trade-off between output diversity and point-forecast accuracy}?

Early likelihood-driven approaches such as TimeGrad~\cite{timegrad} and CSDI~\cite{tashiro2021csdi} maximize log-probability and therefore produce richly diverse samples. However, real-world time series often exhibit pronounced distribution drift between historical and future windows, something the likelihood objective neither detects nor corrects. These models also neglect to make targeted adjustments to the distribution drift of the data, thus often failing to capture the true underlying dynamics of time series. This limitation leads to suboptimal probabilistic performance and fails to provide a solid foundation for accurate point forecasting. As a result, training becomes unstable, sampling variance explodes, and the outputs are “too diverse to be useful,” delivering poor Mean Squared Error (MSE) or Mean Absolute Error (MAE) scores. Consequently, these probabilistic models are seldom compared against strong point-forecasting baselines.

To tame this instability, TimeDiff~\cite{shen2023nonautoregressiveconditionaldiffusionmodels} and mr-Diff~\cite{shen2024multiresolution} prepend a \emph{pre-trained} autoregressive predictor whose outputs serve as the initial trajectory $y_{0:m}$, while directly optimizing performance metrics such as MSE/MAE on the training data. This stitched design stabilizes optimization and improves point prediction accuracy, but it also fixes the diffusion process to a deterministic baseline, constraining the model's inherent ability to explore the full range of possible distributions. As a result, distributional coverage shrinks sharply, and the system reintroduces the maintenance burden of an extra model. These models essentially become similar to regression models, capturing less of the true data distribution.

TMDM~\cite{li2024transformermodulated} goes a step further by jointly training a mature transformer predictor (e.g., Autoformer~\cite{wu2021autoformer} or Non-stationary Transformer~\cite{liu2023nonstationarytransformersexploringstationarity}) and a conditional diffusion model within a Bayesian ELBO framework. This hybrid design mitigates variance and drift, reaching state-of-the-art point metrics among probabilistic forecasters. However, it still relies on an embedded regressor, converges slowly and entails high inference cost.

These observations underscore the need for a \emph{truly simple} diffusion model—one that handles distribution drift, preserves sample diversity, and achieves strong point accuracy \emph{without} auxiliary predictors.
To this end, we propose \textbf{SimDiff}, a simple yet effective diffusion model for time-series forecasting. SimDiff exploits the inherent generative nature of diffusion models to enhance point estimation and remove dependence on pre-trained or jointly trained predictors. Specifically, we address two core questions:

\textbf{1) Can the generative nature of diffusion models be harnessed to enhance point estimation through ensemble methods?} This fundamental question explores how to leverage the generative capabilities of diffusion models to improve point predictions without sacrificing diversity, a gap not effectively addressed in prior work.

\textbf{2) Is it possible to train a purely end-to-end diffusion model without relying on mature regression models to provide the necessary contextual bias?} This inquiry challenges the prevailing reliance on pre-trained or jointly trained external predictors, seeking a simpler yet effective diffusion-based framework.

Our main contributions are summarized as follows:
\begin{itemize}
    \setlength{\itemsep}{1pt}
    \item We propose \textbf{SimDiff}, the first fully end-to-end diffusion model achieving stable SOTA results in time series point forecasting. It employs a unified network as both denoiser and forecaster, greatly simplifying model design.
    \item We introduce \textbf{Normalization Independence (N.I.)}, a diffusion-specific technique that better captures data distributions and mitigates temporal drift.
    \item We design a simple yet efficient transformer backbone, offering clear empirical validation and practical design insights for future studies.
    \item \textbf{SimDiff} matches leading probabilistic models (in CRPS, CRPS-sum) without explicit design by leveraging the inherent generative nature of diffusion. Building on this, we propose a \textbf{Median-of-Means (MoM)} estimator that aggregates probabilistic samples to deliver clear \textbf{SOTA results in point forecasting}. The simplicity of SimDiff further enables much faster inference than existing diffusion-based models, underscoring its superior \textbf{efficiency}.
\end{itemize}

By addressing the aforementioned challenges, SimDiff bridges the gap between probabilistic diversity and accurate point prediction, leveraging the full potential of diffusion models in time series forecasting and offering a practical and powerful framework for future research in this domain.

\begin{figure*}[htbp]
    \centering
    \includegraphics[width=0.9\linewidth]{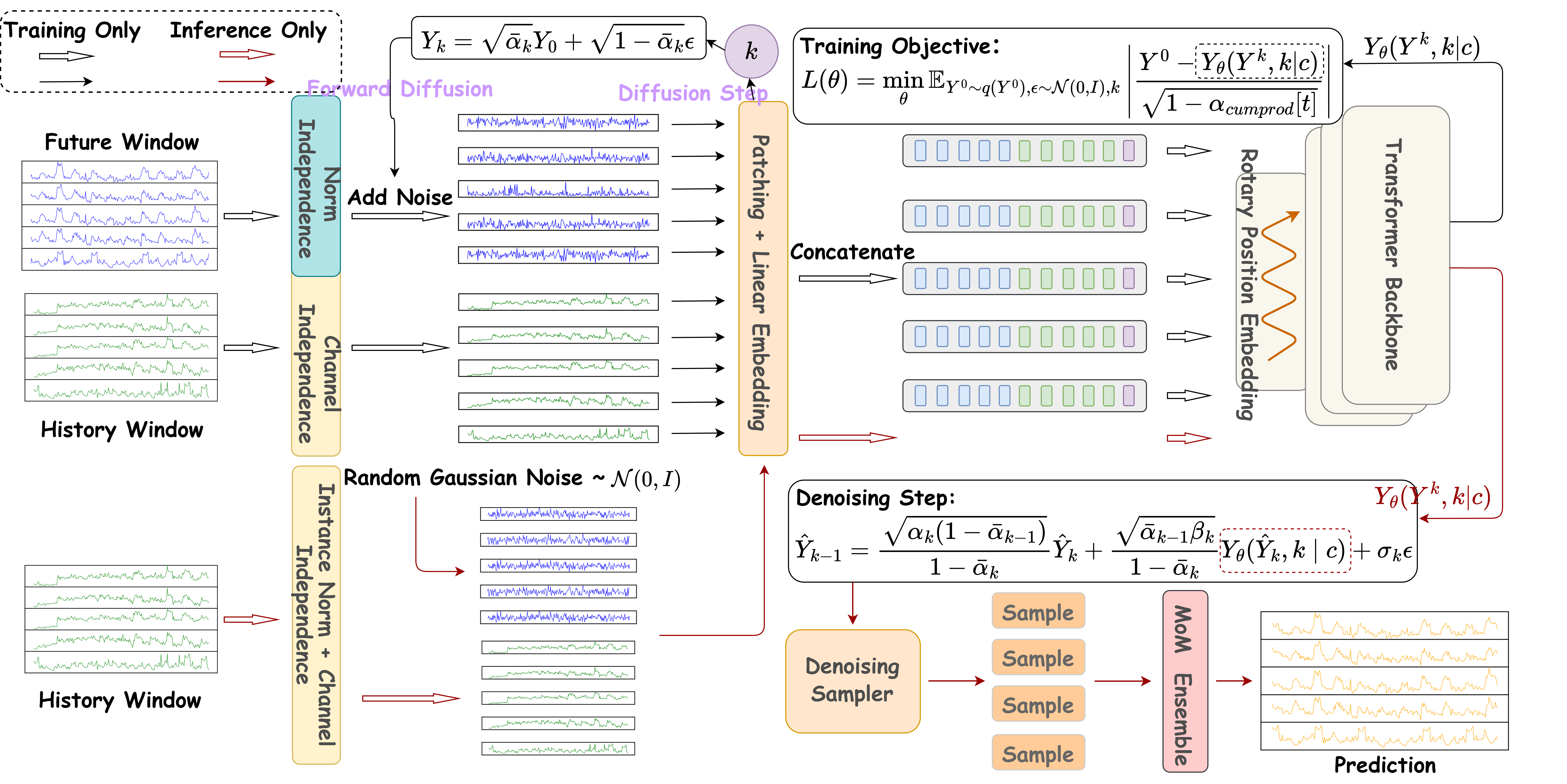}
    \caption{\textbf{SimDiff}: We have developed a streamlined end-to-end patch-based transformer diffusion model for time series forecasting tasks. Key components of our design include Normalization Independence, MoM ensembling, and the incorporation of RoPE.}
    \label{fig:simdiff}
\end{figure*}

\section{SimDiff: Simpler yet Better Diffusion Model }\label{sec:method}

We consider the following problem: given a sequence of multivariate time series observations $\mathbf{X} = x_{-L+1}:x_0$, where each $x_t$ at time step $t$ is a vector of dimension $M$, our goal is to forecast $H$ future values $\mathbf{Y} = x_1:x_H$.
This predictive task is addressed using our proposed \textit{SimDiff} model, which is detailed in Figure \ref{fig:simdiff}. 

\subsection{Diffusion and Denoising For Time Series}
\noindent \textbf{Forward Diffusion Process.} The forward diffusion process for time series forecasting in \textit{SimDiff} follows the methodology of the classical condtional denoising diffusion probabilistic models(DDPMs)~\cite{NEURIPS2020_4c5bcfec}, as is also demonstrated in the former diffusion-TS models ~\cite{tashiro2021csdi, shen2023nonautoregressiveconditionaldiffusionmodels, shen2024multiresolution, timegrad}

In particular, the forward diffusion step for \(Y\) at step \(k\) is given by:
\begin{equation}
Y_k = \sqrt{\bar{\alpha}_k} Y_0 + \sqrt{1 - \bar{\alpha}_k} \epsilon, \quad k = 1, \ldots, K,
\end{equation}
where the noise matrix \(\epsilon\) is sampled from \(\mathcal{N}(0, I)\) with the same size as \(Y\),and \(Y\) is an \(M\)-dimensional vector with a prediction horizon of \(H\).

\noindent \textbf{Backward Denoising Process.} The backward denoising process aims to reconstruct the future time series \(Y\) through a denoising transformer backbone. Each denoising step \(k\) is formulated as:
\begin{align}
\begin{split}
p_{\theta}(Y_{k-1} \mid Y_k, c) 
&= \mathcal{N}\!\left(
    Y_{k-1};\,
    \mu_{\theta}(Y_k, k \mid c),\,
    \sigma_k^2 I
\right),\\
&\quad k = K, \ldots, 1,
\end{split}
\end{align}

where \(\theta\) includes all parameters of the unified conditional denoising transformer, $c$ is a condition derived from the past observations via conditional network, and the mean \(\mu_{\theta}(Y_k, k \mid c)\) is computed as:
{\small
\begin{align}
\mu_{\theta}(Y_k, k \mid c)
= \frac{\sqrt{\alpha_k (1 - \bar{\alpha}_{k-1})}}{1 - \bar{\alpha}_k} Y_k
+ \frac{\sqrt{\bar{\alpha}_{k-1} \beta_k}}{1 - \bar{\alpha}_k}
  Y_{\theta}(Y_k, k \mid c).
\end{align}
}

The denoising objective for learning \(\theta\) is to minimize the gap between the predicted time series and the future ground truth, which will be introduced in detail later.

During inference, the initialization starts from \(\hat{Y}_K \sim \mathcal{N}(0, I)\). For each denoising step \(k\), the update is:
{\small
\begin{align}
\hat{Y}_{k-1}
= \frac{\sqrt{\alpha_k (1 - \bar{\alpha}_{k-1})}}{1 - \bar{\alpha}_k}\, \hat{Y}_k
+ \frac{\sqrt{\bar{\alpha}_{k-1} \beta_k}}{1 - \bar{\alpha}_k}\,
  Y_{\theta}(\hat{Y}_k, k \mid c)
+ \sigma_k \epsilon .
\end{align}
}
where \(\epsilon \sim \mathcal{N}(0, I)\) when \(k > 1\), and \(\epsilon = 0\) otherwise.

By iteratively applying these steps, We can then reconstruct the future time series from the noise.

\begin{figure*}[h]
    \centering
    \begin{subfigure}{0.26\textwidth}
        \centering
        \includegraphics[width=\linewidth]{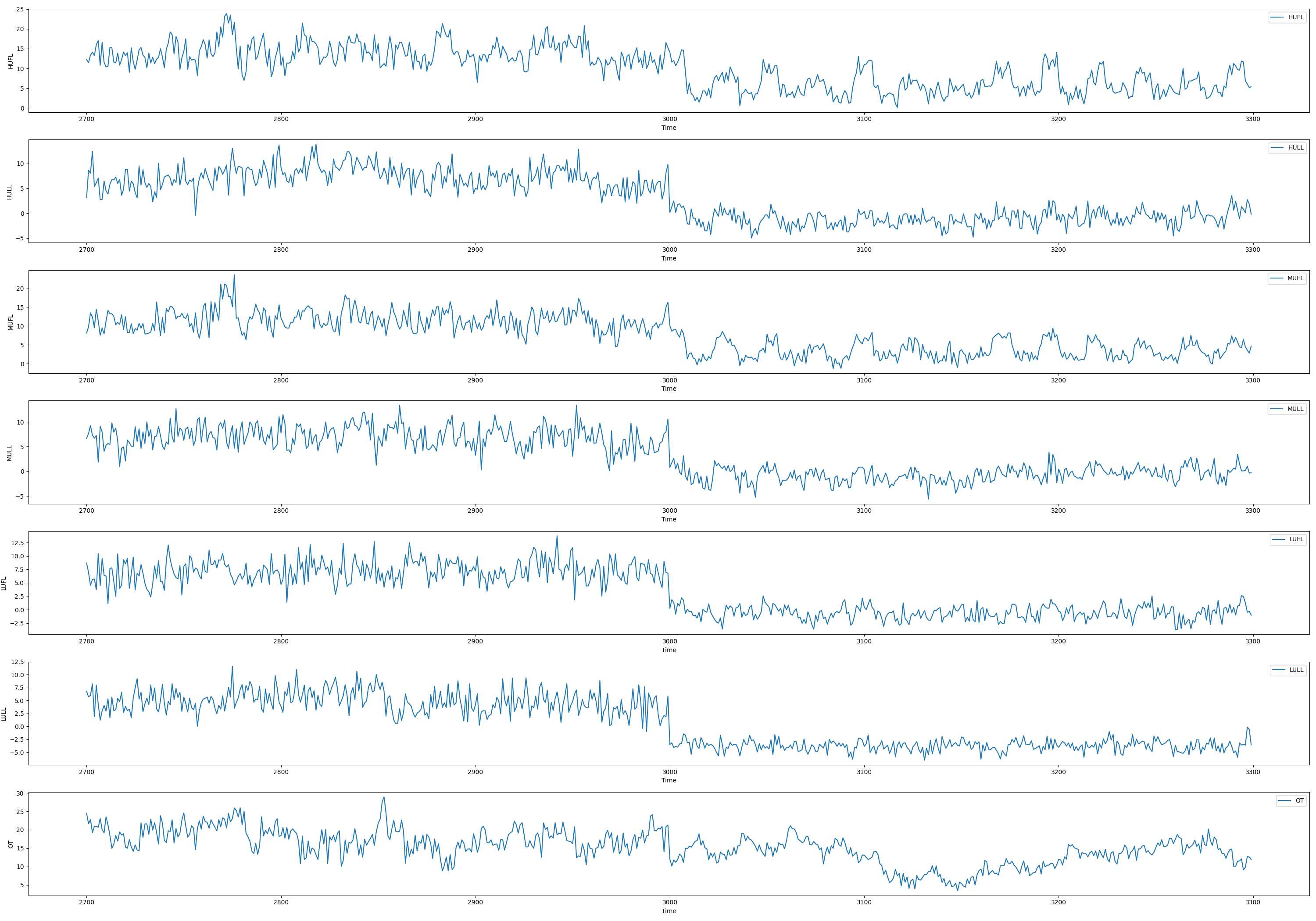}
        \caption{Synthetic Data with Out of Distribution Drift}
        \label{fig:sydata}
    \end{subfigure}
    \hspace{0.05\textwidth}
    \begin{subfigure}{0.26\textwidth}
        \centering
        \includegraphics[width=\linewidth]{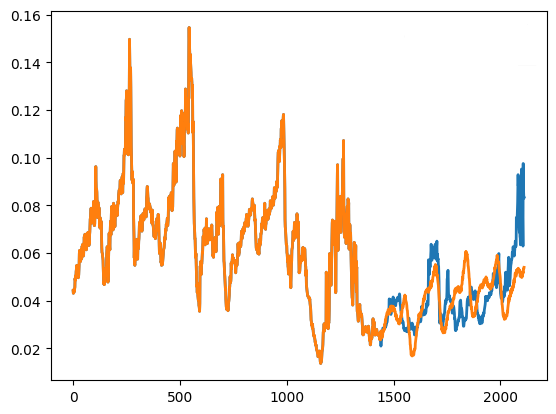}
        \caption{Norm. Independence}
        \label{fig:nidata}
    \end{subfigure}
    \hspace{0.05\textwidth}
    \begin{subfigure}{0.26\textwidth}
        \centering
        \includegraphics[width=\linewidth]{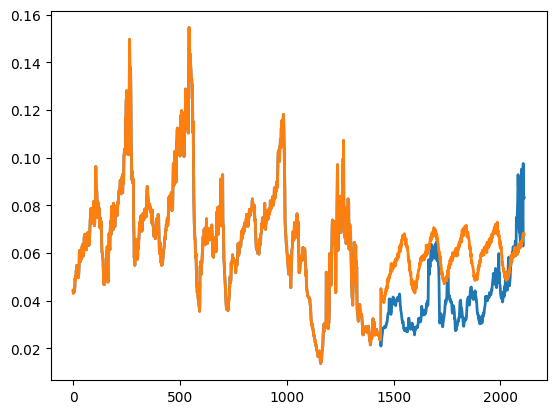}
        \caption{Z-Score Normalization}
        \label{fig:zsdata}
    \end{subfigure}
    \caption{Normalization Independence: Enhancing Robust Training by Mitigating O.O.D.}
    \label{fig:niood}
\end{figure*}

\subsection{Normalization Independence}\label{subsec:NI}
\textbf{Why.}  
Past and future segments of a time series rarely share the same level or scale; normalising both with statistics of the past therefore biases the model when distribution drift occurs. Prior diffusion work nonetheless applies
$\mathbf X_{\!\text{norm}}=(\mathbf X-\mu_{\mathbf X})/\sigma_{\mathbf X}$ and
$\mathbf Y_{\!\text{norm}}=(\mathbf Y-\mu_{\mathbf X})/\sigma_{\mathbf X}$,
implicitly assuming stationarity.

\textbf{What.} 
NI breaks this coupling.  
\emph{Past} samples are instance-normalised and rescaled by learnable $(\boldsymbol\gamma,\boldsymbol\beta)$;  
\emph{future} targets are normalised with their \emph{own} statistics independently \emph{only during training}.  
At test time, the model makes predictions from standard Gaussian noise, then de-normalises them using only past statistics and the learned affine parameters.
This simple change stabilises training, and lets the network learn to infer future scale shifts from the past without the risk of data leakage.

\begin{algorithm}[t]
\begin{algorithmic}[1]
    \Require past $\mathbf{X}$, (future $\mathbf{Y}$ for training), learnable $(\boldsymbol{\gamma},\boldsymbol{\beta})$
    \State Compute $\mu_X$, $\sigma_X$ from $\mathbf{X}$
    \State Normalize $\mathbf{X}$: $\mathbf{X}_{\text{norm}}=\boldsymbol{\gamma}\cdot\dfrac{\mathbf{X}-\mu_X}{\sigma_X}+\boldsymbol{\beta}$
    \If{training}
        \State Compute $\mu_Y$, $\sigma_Y$ from $\mathbf{Y}$
        \State Normalize $\mathbf{Y}$: $\mathbf{Y}_{\text{norm}}=\dfrac{\mathbf{Y}-\mu_Y}{\sigma_Y}$
        \State Corrupt $\mathbf{Y}_{\text{norm}}$ with diffusion noise; optimize loss
    \Else
        \State Sample standard Gaussian noise $\boldsymbol{\epsilon}$
        \State Cond-DDPM denoising on $\boldsymbol{\epsilon}$ conditioned on $\mathbf{X}_{\text{norm}}$, producing $\widehat{\mathbf{Y}}_{\text{norm}}$
        \State De-normalize: $\widehat{\mathbf{Y}}=\sigma_X\cdot\dfrac{\widehat{\mathbf{Y}}_{\text{norm}}-\boldsymbol{\beta}}{\boldsymbol{\gamma}}+\mu_X$
    \EndIf
\end{algorithmic}
\caption{N.I. in training vs.\ inference}
\label{alg:ni-alg}
\end{algorithm}

NI adds only a lightweight affine layer with negligible cost, but markedly improves robustness to distribution drift by better aligning training data with the diffusion's Gaussian prior (Fig.~\ref{fig:niood}). Detailed ablations can be found in the later section.

\subsection{Transformer Denoising Network}

Our \textit{SimDiff} model uses a transformer-based denoising network specifically designed for time series forecasting, moving away from complex stacked CNNs or U-Nets. This architecture leverages transformer's strengths in capturing temporal dependencies, essential for unstationary time series analysis.

\noindent \textbf{Patch-based Tokenization.} We utilize patching~\cite{Yuqietal-2023-PatchTST,zhang2023crossformer} to convert time series into overlapping tokens, with each patch acting as a token for local dependencies. A dense MLP transforms these patches into token embeddings, and diffusion timesteps are similarly processed into a time token, which is concatenated with the original tokens.

\noindent \textbf{Rotary Position Embedding (RoPE).}To better capture temporal order in long-term forecasting, we employ Rotary Position Embedding (RoPE)~\cite{su2023roformerenhancedtransformerrotary}. By encoding relative positional information through rotational transformations, RoPE preserves temporal dependencies and strengthens the attention mechanism’s ability to focus across time, enhancing modeling of dynamic patterns.

\noindent \textbf{Channel Independence and No Skip Connections.} Skip connections, as used in U-ViT~\cite{bao2022all}, help preserve spatial features but can amplify noise in time series, distorting diffusion distributions and degrading performance. To mitigate this, \textit{SimDiff} removes skip connections for more stable modeling. Additionally, it employs channel independence~\cite{Yuqietal-2023-PatchTST}, processing each channel separately to enhance efficiency and reduce complexity. This increases data volume, improves distribution learning, and enables global attention to focus on essential temporal patterns for more accurate forecasting.

These designs above enable \textit{SimDiff} to balance simplicity and depth, ensuring robust and efficient time series prediction. More analysis on design choices can be found in the supplementary materials.

\begin{figure}[t]
    \centering
    \includegraphics[width=.8\linewidth]{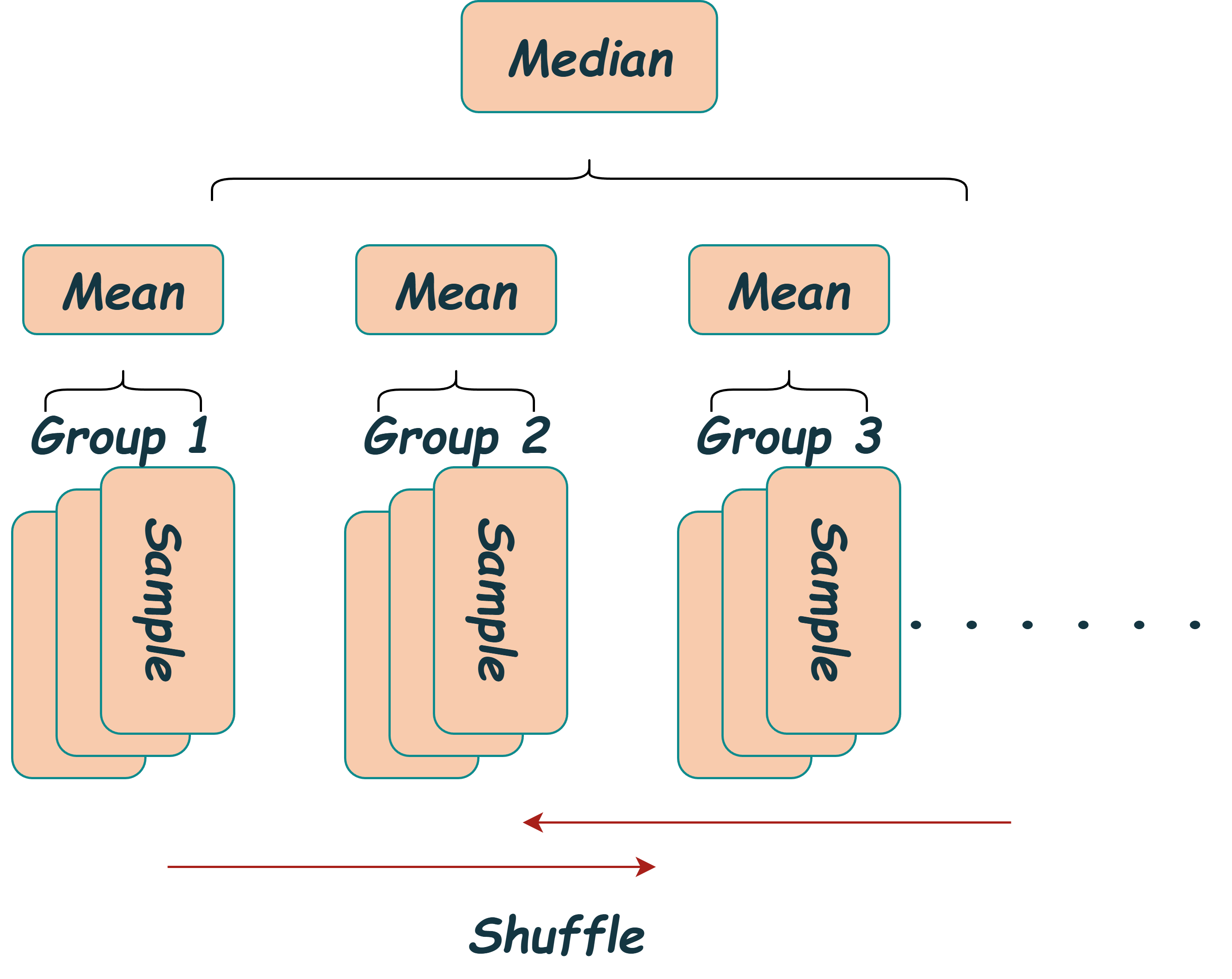}
    \caption{The MoM Ensemble}
    \label{fig:mom}
\end{figure}

\subsection{Median of Means Ensemble}

Diffusion models inherently explore a wide range of possible probability traces, where extreme values are often unavoidable. To mitigate the influence of these outliers while still faithfully capturing the overall distribution trend, we intuitively need a proper strategy to derive a stable final estimate from the sampled probabilities. Therefore, we introduce the MoM estimator to transition from efficient distribution prediction to precise point estimation.

Originally a statistical method, we reintroduced MoM in our model as a reliable approach to estimate true values from multiple probabilistic samples. The MoM estimator divides a dataset of size \( n \) into \( K \) subsamples of size \( B \), computes their means \(\hat{\mu}_1, \ldots, \hat{\mu}_K\), and takes the median. To improve robustness, this process is repeated \( R \) times with shuffled data. The final estimator is the average of the \( R \) medians (just as the Figure \ref{fig:mom} shows):
\begin{align}
\hat{\mu}_{\text{MoM}} = \frac{1}{R} \sum_{r=1}^{R} \text{median}(\hat{\mu}_{1}^{(r)}, \ldots, \hat{\mu}_{K}^{(r)})
\end{align}

\subsection{Loss Function for Robust Training}
Different from the literature, we choose a weighted mean absolute error (MAE) loss as the denoising objective.The loss function is expressed as:
\begin{align}
L(\theta) = \min_{\theta} \mathbb{E}_{Y^0 \sim q(Y^0),\epsilon \sim \mathcal{N}(0, I),k} \left| \frac{Y^0 - Y_{\theta}(Y^k, k | c)}{\sqrt{1 - \alpha_{\textit{cumprod}}[k]}} \right|
\end{align}
where \( Y_0 \) represents the target values at the initial timestep, \( Y(\theta; k, c) \) denotes the model output,and \(\alpha_{\text{cumprod}}[t]\) refers to the cumulative product of \(1 - \alpha\) up to timestep \( t \), which adjusts the normalization factor to account for the accumulated noise reduction over the diffusion steps.This scaling is critical as it allows the model to focus learning on periods with higher noise levels, ensuring robustness and accuracy in the denoising performance across varying conditions in the diffusion process. 

\section{Experiments}

\subsection{Probabilistic Forecasting Underpins Accurate Point Estimates}

Diffusion models excel at modelling predictive distributions, a property that naturally supports accurate point forecasts. Estimating a probability distribution is inherently more challenging than predicting a single point, as it requires capturing both central tendencies and distributional boundaries. Effective point prediction thus indicates how well the model understands these distributional margins—a model capable of accurate distribution modelling is also more likely to produce precise point predictions. To quantitatively evaluate distributional quality, we primarily rely on the Continuous Ranked Probability Score (CRPS) and its aggregate variant, CRPS-sum, because they measure the full distance between predicted and empirical distributions.

\begin{table}[h]
\centering
\resizebox{\linewidth}{!}{
\begin{tabular}{!{\vrule width 0pt}l|cc|cc|cc|cc!{\vrule width 0pt}}
\toprule
\multirow{2}{*}{Method} & \multicolumn{2}{c|}{\textit{ELEC.}} & \multicolumn{2}{c|}{\textit{TRAFFIC}} & \multicolumn{2}{c|}{\textit{TAXI}} & \multicolumn{2}{c}{\textit{WIKI.}} \\
& C. & C.S. & C. & C.S. & C. & C.S. & C. & C.S. \\
\midrule
GP-Copula    & 0.77 & 0.024 & 0.48 & 0.078 & 0.54 & 0.208 & 0.71 & 0.086 \\
LSRP         & 0.45 & 0.024 & 0.43 & 0.078 & 0.49 & 0.175 & 0.59 & 0.078 \\
LSTMMAF      & 0.41 & 0.023 & 0.45 & 0.069 & 0.46 & 0.161 & 0.55 & 0.067 \\
\midrule
CSDI         & 0.37 & 0.029 & 0.32 & 0.053 & -     & -     & -     & -     \\
D3VAE        & 0.33 & 0.030 & 0.29 & 0.049 & \textbf{0.35} & \textbf{0.130} & 0.52 & 0.069 \\
LDT          & 0.27 & 0.021 & 0.23 & \underline{0.040} & \underline{0.36} & 0.131 & 0.46 & 0.063 \\
TMDM         &  \underline{0.24} & 0.023 &\underline{0.21} & 0.042 & 0.44 & 0.172 & \underline{0.43} & \underline{0.060} \\
MG-TSD     & 0.25 & 0.023 & 0.38 & 0.044 & \underline{0.36} &\textbf{0.130} & 0.46 & 0.063 \\
TSDiff     & \underline{0.24}    & \underline{0.020} & 0.34 & 0.046 & 0.40 & 0.155 & 0.45 & 0.066 \\
TimeGrad     & 0.34 & 0.025 & 0.35 & 0.050 & 0.40 & 0.137 & 0.55 & 0.064 \\
SSSD         & 0.35 & 0.026 & 0.33 & 0.047 & 0.39 & 0.133 & 0.53 & 0.065 \\
\midrule
\textit{SimDiff} & \textbf{0.22} & \textbf{0.019} & \textbf{0.16} & \textbf{0.039} & 0.42 & 0.166 & \textbf{0.41} & \textbf{0.057} \\
\bottomrule
\end{tabular}
}
\caption{Testing CRPS and CRPS-Sum in the multivariate setting. Best: \textbf{bold}, the second best: \underline{underlined}. - denotes out of GPU memory.}
\label{tab:probb}
\end{table}

\textbf{Results.} Table~\ref{tab:probb} demonstrates that SimDiff achieves strong probabilistic forecasting performance across datasets despite not being explicitly optimized for this probabilistic task. Instead, it leverages the inherent generative capacity of diffusion through a single, carefully designed diffusion–Transformer framework. This effective distributional fit establishes a solid foundation for the point forecasting gains discussed in Section~\ref{sec:main_results}. Compared to baselines, SimDiff robustly adapts to diverse series, effectively balancing prediction diversity and precision under dynamic distribution shifts.

In summary, SimDiff’s robust probabilistic performance supports its SOTA point forecasting accuracy, illustrating that reliable distribution modelling and precise point estimation can coexist within a simple, end-to-end diffusion framework.

\subsection{Accurate Long-Term Time Series Point Forecasting}\label{sec:main_results}
\begin{table*}[h]
\centering
\resizebox{0.9\textwidth}{!}{%
\begin{tabular}{c|lllllllll|c}%
\specialrule{1.1pt}{0pt}{0pt}
\ Method & \textit{ NorPool} & \textit{ Caiso} & \textit{ Traffic} & \textit{ Electricity} & \textit{ Weather} & \textit{ Exchange} & \textit{ ETTh1} & \textit{ ETTm1} & \textit{ Wind} & Rank \\ 
\midrule
\textbf{OURS} & $\textbf{0.534}_{\raisebox{0.5ex}{\footnotesize (1)}}$ & $\underline{0.106}_{\raisebox{0.5ex}{\footnotesize (2)}}$ & $\underline{0.383}_{\raisebox{0.5ex}{\footnotesize (2)}}$ & $\textbf{0.145}_{\raisebox{0.5ex}{\footnotesize (1)}}$ & $\underline{0.299}_{\raisebox{0.5ex}{\footnotesize (2)}}$ & $\textbf{0.015}_{\raisebox{0.5ex}{\footnotesize (1)}}$ & $\textbf{0.394}_{\raisebox{0.5ex}{\footnotesize (1)}}$ & $\textbf{0.322}_{\raisebox{0.5ex}{\footnotesize (1)}}$ & $\textbf{0.880}_{\raisebox{0.5ex}{\footnotesize (1)}}$ & $\textbf{1.33}$\\
\midrule
mr-Diff & $0.645_{\raisebox{0.5ex}{\footnotesize (4)}}$ & $0.127_{\raisebox{0.5ex}{\footnotesize (5)}}$ & $0.474_{\raisebox{0.5ex}{\footnotesize (8)}}$ & $0.155_{\raisebox{0.5ex}{\footnotesize (5)}}$ & $\textbf{0.296}_{\raisebox{0.5ex}{\footnotesize (1)}}$ & $\underline{0.016}_{\raisebox{0.5ex}{\footnotesize (2)}}$ & $0.411_{\raisebox{0.5ex}{\footnotesize (5)}}$ & $0.340_{\raisebox{0.5ex}{\footnotesize (4)}}$ & $\underline{0.881}_{\raisebox{0.5ex}{\footnotesize (2)}}$ & $4.00$ \\
TimeDiff & $0.665_{\raisebox{0.5ex}{\footnotesize (6)}}$ & $0.136_{\raisebox{0.5ex}{\footnotesize (8)}}$ & $0.564_{\raisebox{0.5ex}{\footnotesize (10)}}$ & $0.193_{\raisebox{0.5ex}{\footnotesize (7)}}$ & $0.311_{\raisebox{0.5ex}{\footnotesize (4)}}$ & $0.018_{\raisebox{0.5ex}{\footnotesize (8)}}$ & $0.407_{\raisebox{0.5ex}{\footnotesize (3)}}$ & $\underline{0.336}_{\raisebox{0.5ex}{\footnotesize (2)}}$ & $0.896_{\raisebox{0.5ex}{\footnotesize (3)}}$ &$5.67$ \\
TimeGrad & $1.152_{\raisebox{0.5ex}{\footnotesize (22)}}$ & $0.258_{\raisebox{0.5ex}{\footnotesize (20)}}$ & $1.745_{\raisebox{0.5ex}{\footnotesize (24)}}$ & $0.736_{\raisebox{0.5ex}{\footnotesize (23)}}$ & $0.392_{\raisebox{0.5ex}{\footnotesize (16)}}$ & $0.079_{\raisebox{0.5ex}{\footnotesize (22)}}$ & $0.993_{\raisebox{0.5ex}{\footnotesize (24)}}$ & $0.874_{\raisebox{0.5ex}{\footnotesize (23)}}$ & $1.209_{\raisebox{0.5ex}{\footnotesize (23)}}$ &$21.89$ \\
TMDM & $0.681_{\raisebox{0.5ex}{\footnotesize (8)}}$ &$0.214_{\raisebox{0.5ex}{\footnotesize (14)}}$ & $0.513_{\raisebox{0.5ex}{\footnotesize (9)}}$ & $0.267_{\raisebox{0.5ex}{\footnotesize (14)}}$ &$0.403_{\raisebox{0.5ex}{\footnotesize (18)}}$  &$0.023_{\raisebox{0.5ex}{\footnotesize (13)}}$  &$0.535_{\raisebox{0.5ex}{\footnotesize (13)}}$ &$0.436_{\raisebox{0.5ex}{\footnotesize (14)}}$ &$0.901_{\raisebox{0.5ex}{\footnotesize (5)}}$ &$12.00$\\
CSDI & $1.011_{\raisebox{0.5ex}{\footnotesize (21)}}$ & $0.253_{\raisebox{0.5ex}{\footnotesize (19)}}$ & - & - & $0.356_{\raisebox{0.5ex}{\footnotesize (11)}}$ & $0.077_{\raisebox{0.5ex}{\footnotesize (21)}}$ & $0.497_{\raisebox{0.5ex}{\footnotesize (9)}}$ & $0.529_{\raisebox{0.5ex}{\footnotesize (19)}}$ & $1.066_{\raisebox{0.5ex}{\footnotesize (12)}}$ &$16.00$\\
SSSD & $0.872_{\raisebox{0.5ex}{\footnotesize (14)}}$ & $0.195_{\raisebox{0.5ex}{\footnotesize (11)}}$ & $0.642_{\raisebox{0.5ex}{\footnotesize (13)}}$ & $0.255_{\raisebox{0.5ex}{\footnotesize (13)}}$ & $0.349_{\raisebox{0.5ex}{\footnotesize (10)}}$ & $0.061_{\raisebox{0.5ex}{\footnotesize (18)}}$ & $0.726_{\raisebox{0.5ex}{\footnotesize (20)}}$ & $0.464_{\raisebox{0.5ex}{\footnotesize (15)}}$ & $1.188_{\raisebox{0.5ex}{\footnotesize (21)}}$ &$15.00$\\
\midrule
D3VAE & $0.745_{\raisebox{0.5ex}{\footnotesize (12)}}$ & $0.241_{\raisebox{0.5ex}{\footnotesize (18)}}$ & $0.928_{\raisebox{0.5ex}{\footnotesize (19)}}$ & $0.286_{\raisebox{0.5ex}{\footnotesize (17)}}$ & $0.375_{\raisebox{0.5ex}{\footnotesize (13)}}$ & $0.200_{\raisebox{0.5ex}{\footnotesize (24)}}$ & $0.504_{\raisebox{0.5ex}{\footnotesize (11)}}$ & $0.362_{\raisebox{0.5ex}{\footnotesize (10)}}$ & $1.118_{\raisebox{0.5ex}{\footnotesize (17)}}$ &$15.67$\\
CPF & $1.613_{\raisebox{0.5ex}{\footnotesize (25)}}$ & $0.383_{\raisebox{0.5ex}{\footnotesize (22)}}$ & $1.625_{\raisebox{0.5ex}{\footnotesize (23)}}$ & $0.793_{\raisebox{0.5ex}{\footnotesize (24)}}$ & $1.390_{\raisebox{0.5ex}{\footnotesize (25)}}$ & $\underline{0.016}_{\raisebox{0.5ex}{\footnotesize (2)}}$ & $0.730_{\raisebox{0.5ex}{\footnotesize (21)}}$ & $0.482_{\raisebox{0.5ex}{\footnotesize (17)}}$ & $1.140_{\raisebox{0.5ex}{\footnotesize (19)}}$ &$19.78$\\
PSA-GAN & $1.501_{\raisebox{0.5ex}{\footnotesize (24)}}$ & $0.510_{\raisebox{0.5ex}{\footnotesize (24)}}$ & $1.614_{\raisebox{0.5ex}{\footnotesize (22)}}$ & $0.535_{\raisebox{0.5ex}{\footnotesize (22)}}$ & $1.220_{\raisebox{0.5ex}{\footnotesize (23)}}$ & $0.018_{\raisebox{0.5ex}{\footnotesize (8)}}$ & $0.623_{\raisebox{0.5ex}{\footnotesize (19)}}$ & $0.537_{\raisebox{0.5ex}{\footnotesize (20)}}$ & $1.127_{\raisebox{0.5ex}{\footnotesize (18)}}$ &$20.00$\\
\midrule
N-Hits & $0.716_{\raisebox{0.5ex}{\footnotesize (10)}}$ & $0.131_{\raisebox{0.5ex}{\footnotesize (6)}}$ & $0.386_{\raisebox{0.5ex}{\footnotesize (4)}}$ & $0.152_{\raisebox{0.5ex}{\footnotesize (4)}}$ & $0.323_{\raisebox{0.5ex}{\footnotesize (6)}}$ & $0.017_{\raisebox{0.5ex}{\footnotesize (7)}}$ & $0.498_{\raisebox{0.5ex}{\footnotesize (10)}}$ & $0.353_{\raisebox{0.5ex}{\footnotesize (8)}}$ & $1.033_{\raisebox{0.5ex}{\footnotesize (9)}}$ &$7.11$\\
FiLM & $0.723_{\raisebox{0.5ex}{\footnotesize (11)}}$ & $0.179_{\raisebox{0.5ex}{\footnotesize (10)}}$ & $0.628_{\raisebox{0.5ex}{\footnotesize (13)}}$ & $0.210_{\raisebox{0.5ex}{\footnotesize (10)}}$ & $0.327_{\raisebox{0.5ex}{\footnotesize (7)}}$ & $\underline{0.016}_{\raisebox{0.5ex}{\footnotesize (2)}}$ & $0.426_{\raisebox{0.5ex}{\footnotesize (7)}}$ & $0.347_{\raisebox{0.5ex}{\footnotesize (6)}}$ & $0.984_{\raisebox{0.5ex}{\footnotesize (6)}}$ &$8.00$\\
Depts & $0.662_{\raisebox{0.5ex}{\footnotesize (5)}}$ & $\underline{0.106}_{\raisebox{0.5ex}{\footnotesize (2)}}$ & $1.019_{\raisebox{0.5ex}{\footnotesize (21)}}$ & $0.319_{\raisebox{0.5ex}{\footnotesize (19)}}$ & $0.761_{\raisebox{0.5ex}{\footnotesize (22)}}$ & $0.020_{\raisebox{0.5ex}{\footnotesize (11)}}$ & $0.579_{\raisebox{0.5ex}{\footnotesize (15)}}$ & $0.380_{\raisebox{0.5ex}{\footnotesize (11)}}$ & $1.082_{\raisebox{0.5ex}{\footnotesize (14)}}$ &$13.33$\\
NBeats & $0.832_{\raisebox{0.5ex}{\footnotesize (13)}}$ & $0.141_{\raisebox{0.5ex}{\footnotesize (9)}}$ & $\textbf{0.373}_{\raisebox{0.5ex}{\footnotesize (1)}}$ & $0.269_{\raisebox{0.5ex}{\footnotesize (15)}}$ & $1.344_{\raisebox{0.5ex}{\footnotesize (24)}}$ & $\underline{0.016}_{\raisebox{0.5ex}{\footnotesize (2)}}$ & $0.586_{\raisebox{0.5ex}{\footnotesize (17)}}$ & $0.391_{\raisebox{0.5ex}{\footnotesize (12)}}$ & $1.069_{\raisebox{0.5ex}{\footnotesize (13)}}$ &$11.78$\\
\midrule
Scaleformer & $0.983_{\raisebox{0.5ex}{\footnotesize (17)}}$ & $0.207_{\raisebox{0.5ex}{\footnotesize (13)}}$ & $0.618_{\raisebox{0.5ex}{\footnotesize (12)}}$ & $0.195_{\raisebox{0.5ex}{\footnotesize (8)}}$ & $0.462_{\raisebox{0.5ex}{\footnotesize (19)}}$ & $0.036_{\raisebox{0.5ex}{\footnotesize (15)}}$ & $0.613_{\raisebox{0.5ex}{\footnotesize (18)}}$ & $0.481_{\raisebox{0.5ex}{\footnotesize (16)}}$ & $1.359_{\raisebox{0.5ex}{\footnotesize (24)}}$ &$15.78$\\
PatchTST & $\underline{0.547}_{\raisebox{0.5ex}{\footnotesize (2)}}$ & $0.110_{\raisebox{0.5ex}{\footnotesize (4)}}$ & $0.385_{\raisebox{0.5ex}{\footnotesize (3)}}$ & $\underline{0.147}_{\raisebox{0.5ex}{\footnotesize (2)}}$ & $0.302_{\raisebox{0.5ex}{\footnotesize (3)}}$ & $\underline{0.016}_{\raisebox{0.5ex}{\footnotesize (2)}}$ & $\underline{0.405}_{\raisebox{0.5ex}{\footnotesize (2)}}$ & $0.337_{\raisebox{0.5ex}{\footnotesize (3)}}$ & $1.017_{\raisebox{0.5ex}{\footnotesize (8)}}$ &$\underline{3.22}$\\
FedFormer & $0.873_{\raisebox{0.5ex}{\footnotesize (15)}}$ & $0.205_{\raisebox{0.5ex}{\footnotesize (12)}}$ & $0.591_{\raisebox{0.5ex}{\footnotesize (11)}}$ & $0.238_{\raisebox{0.5ex}{\footnotesize (12)}}$ & $0.342_{\raisebox{0.5ex}{\footnotesize (9)}}$ & $0.133_{\raisebox{0.5ex}{\footnotesize (23)}}$ & $0.541_{\raisebox{0.5ex}{\footnotesize (14)}}$ & $0.426_{\raisebox{0.5ex}{\footnotesize (13)}}$ & $1.113_{\raisebox{0.5ex}{\footnotesize (16)}}$ &$13.89$\\
Autoformer & $0.940_{\raisebox{0.5ex}{\footnotesize (16)}}$ & $0.226_{\raisebox{0.5ex}{\footnotesize (16)}}$ & $0.688_{\raisebox{0.5ex}{\footnotesize (18)}}$ & $0.201_{\raisebox{0.5ex}{\footnotesize (9)}}$ & $0.360_{\raisebox{0.5ex}{\footnotesize (12)}}$ & $0.056_{\raisebox{0.5ex}{\footnotesize (17)}}$ & $0.516_{\raisebox{0.5ex}{\footnotesize (12)}}$ & $0.565_{\raisebox{0.5ex}{\footnotesize (21)}}$ & $1.083_{\raisebox{0.5ex}{\footnotesize (15)}}$ &$15.11$\\
Pyraformer & $1.008_{\raisebox{0.5ex}{\footnotesize (20)}}$ & $0.273_{\raisebox{0.5ex}{\footnotesize (21)}}$ & $0.659_{\raisebox{0.5ex}{\footnotesize (15)}}$ & $0.273_{\raisebox{0.5ex}{\footnotesize (16)}}$ & $0.394_{\raisebox{0.5ex}{\footnotesize (17)}}$ & $0.032_{\raisebox{0.5ex}{\footnotesize (14)}}$ & $0.579_{\raisebox{0.5ex}{\footnotesize (15)}}$ & $0.493_{\raisebox{0.5ex}{\footnotesize (18)}}$ & $1.061_{\raisebox{0.5ex}{\footnotesize (11)}}$ &$16.33$\\
Informer & $0.985_{\raisebox{0.5ex}{\footnotesize (18)}}$ & $0.231_{\raisebox{0.5ex}{\footnotesize (17)}}$ & $0.664_{\raisebox{0.5ex}{\footnotesize (16)}}$ & $0.298_{\raisebox{0.5ex}{\footnotesize (18)}}$ & $0.385_{\raisebox{0.5ex}{\footnotesize (14)}}$ & $0.073_{\raisebox{0.5ex}{\footnotesize (20)}}$ & $0.775_{\raisebox{0.5ex}{\footnotesize (23)}}$ & $0.673_{\raisebox{0.5ex}{\footnotesize (22)}}$ & $1.168_{\raisebox{0.5ex}{\footnotesize (20)}}$ &$18.67$\\
Transformer & $1.005_{\raisebox{0.5ex}{\footnotesize (19)}}$ & $0.206_{\raisebox{0.5ex}{\footnotesize (13)}}$ & $0.671_{\raisebox{0.5ex}{\footnotesize (17)}}$ & $0.328_{\raisebox{0.5ex}{\footnotesize (20)}}$ & $0.388_{\raisebox{0.5ex}{\footnotesize (15)}}$ & $0.062_{\raisebox{0.5ex}{\footnotesize (19)}}$ & $0.759_{\raisebox{0.5ex}{\footnotesize (22)}}$ & $0.992_{\raisebox{0.5ex}{\footnotesize (24)}}$ & $1.201_{\raisebox{0.5ex}{\footnotesize (22)}}$ &$19.00$\\
\midrule
SCINet & $0.613_{\raisebox{0.5ex}{\footnotesize (3)}}$ & $\textbf{0.095}_{\raisebox{0.5ex}{\footnotesize (1)}}$ & $0.434_{\raisebox{0.5ex}{\footnotesize (7)}}$ & $0.171_{\raisebox{0.5ex}{\footnotesize (6)}}$ & $0.329_{\raisebox{0.5ex}{\footnotesize (8)}}$ & $0.036_{\raisebox{0.5ex}{\footnotesize (15)}}$ & $0.465_{\raisebox{0.5ex}{\footnotesize (8)}}$ & $0.359_{\raisebox{0.5ex}{\footnotesize (9)}}$ & $1.055_{\raisebox{0.5ex}{\footnotesize (10)}}$ &$7.44$\\
NLinear & $0.707_{\raisebox{0.5ex}{\footnotesize (9)}}$ & $0.135_{\raisebox{0.5ex}{\footnotesize (7)}}$ & $0.430_{\raisebox{0.5ex}{\footnotesize (6)}}$ & $\underline{0.147}_{\raisebox{0.5ex}{\footnotesize (2)}}$ & $0.313_{\raisebox{0.5ex}{\footnotesize (5)}}$ & $0.019_{\raisebox{0.5ex}{\footnotesize (10)}}$ & $0.410_{\raisebox{0.5ex}{\footnotesize (4)}}$ & $0.349_{\raisebox{0.5ex}{\footnotesize (7)}}$ & $0.989_{\raisebox{0.5ex}{\footnotesize (7)}}$ &$6.33$\\
DLinear & $0.670_{\raisebox{0.5ex}{\footnotesize (7)}}$ & $0.461_{\raisebox{0.5ex}{\footnotesize (23)}}$ & $0.389_{\raisebox{0.5ex}{\footnotesize (5)}}$ & $0.215_{\raisebox{0.5ex}{\footnotesize (11)}}$ & $0.488_{\raisebox{0.5ex}{\footnotesize (20)}}$ & $0.022_{\raisebox{0.5ex}{\footnotesize (12)}}$ & $0.415_{\raisebox{0.5ex}{\footnotesize (6)}}$ & $0.345_{\raisebox{0.5ex}{\footnotesize (5)}}$ & $0.899_{\raisebox{0.5ex}{\footnotesize (4)}}$ &$10.33$\\
LSTMa & $1.481_{\raisebox{0.5ex}{\footnotesize (23)}}$ & $0.217_{\raisebox{0.5ex}{\footnotesize (16)}}$ & $0.966_{\raisebox{0.5ex}{\footnotesize (20)}}$ & $0.414_{\raisebox{0.5ex}{\footnotesize (21)}}$ & $0.662_{\raisebox{0.5ex}{\footnotesize (21)}}$ & $0.403_{\raisebox{0.5ex}{\footnotesize (25)}}$ & $1.149_{\raisebox{0.5ex}{\footnotesize (25)}}$ & $1.030_{\raisebox{0.5ex}{\footnotesize (25)}}$ & $1.464_{\raisebox{0.5ex}{\footnotesize (25)}}$ &$22.33$\\
\specialrule{1.1pt}{0pt}{0pt}
\end{tabular}
}
\caption{Testing MSE in multivariate settings. The number in brackets indicates the rank. Best: \textbf{bold}, the second best: \underline{underlined}. - denotes out of GPU memory. The results are the average of 5 runs. Our results are stable with a variance of 5 runs less than 1e-5.}
\label{tab:main_mse}
\end{table*}

Building on this strong distributional grasp, our Median of Means (MoM) estimator effectively translates probabilistic forecasts into robust and precise point predictions.

\noindent \textbf{Results.} Table \ref{tab:main_mse} presents the MSE in the multivariate setting, where the proposed \textit{SimDiff} model achieves the best performance on 6 out of 9 datasets, with particularly notable improvements on challenging datasets such as Norpool and ETTh1. Even on the remaining 3 datasets, \textit{SimDiff} secures the second-best rankings, demonstrating its robustness across diverse data types. On large and complex datasets like Traffic and Electricity, where other diffusion models tend to underperform, \textit{SimDiff} also achieves state-of-the-art or comparable results. Quantitatively, \textit{SimDiff} reduces the MSE by an average of \textbf{8.3\%} across all datasets compared to other diffusion models like mr-Diff, showing substantial improvements. These results underscore \textit{SimDiff}'s consistent superiority over other competitive models, as reflected in its rankings. The full experiment settings can be found in the appendix material.

\subsection{Standard Diffusion Transformer Suffices for Accurate Point Forecasting}
\begin{table*}[h]
\centering
\resizebox{0.8\textwidth}{!}{
\begin{tabular}{c|ccc|ccc|ccc|ccc}
\toprule
\textbf{Models}& \multicolumn{3}{c|}{\textit{$SimDiff$}\tiny{ (1 Stage, Point)}} & \multicolumn{3}{c|}{\textit{$TimeDiff$}\tiny{ (2 Stage, Point)}}& \multicolumn{3}{c|}{\textit{$CSDI$}\tiny{ (1 Stage, Prob)}} & \multicolumn{3}{c}{\textit{$TimeGrad$}\tiny{ (1 Stage, Prob)}} \\
\midrule
\textbf{Datasets}& \textit{$MSE$} & \textit{$MSE_E$} & \textit{$Var.$} & \textit{$MSE$} & \textit{$MSE_E$} & \textit{$Var.$} & \textit{$MSE$} & \textit{$MSE_E$} & \textit{$Var.$} & \textit{$MSE$} & \textit{$MSE_E$} & \textit{$Var.$} \\
\midrule
$ETTh1$ & $0.408$  & $\underline{0.394}$  & $\textbf{0.012}$  &  $0.407$ & $\underline{0.405}$ &   $0.00081$  & $0.497$  & $\underline{0.494}$  & $0.017$& $0.993$  & $\underline{0.990}$  & $0.013$\\
$ETTm1$ & $0.333$  &  $\underline{0.322}$ &  $\textbf{0.012}$ &  $\underline{0.336}$ & $0.342$ &   $0.00072$ & $0.529$  & $\underline{0.527}$  & $0.039$ & $\underline{0.874}$  & $0.891$  & $0.061$\\
$Weather$ &  $0.317$ &  $\underline{0.299}$ & $\textbf{0.005}$  & $0.311$  & $\underline{0.309}$ &  $0.00064$  & $\underline{0.356}$  & $0.359$  & $0.029$& $\underline{0.392}$  & $0.393$  & $0.034$\\
$NorPool$ &  $0.548$ &  $\underline{0.534}$ & $\textbf{0.011}$  & $\underline{0.665}$  & $0.670$ &  $0.00062$  & $\underline{1.011}$  & $1.210$ &  $0.084$ & $\underline{1.152}$  & $1.189$  & $0.085$\\
$Wind$ &  $0.901$ &  $\underline{0.880}$ &  $\textbf{0.018}$ &  $0.896$ & $\underline{0.891}$ &   $0.00074$ & $1.066$  & $\underline{1.013}$ &  $0.012$ & $1.209$  & $\underline{1.201}$  & $0.037$\\
\bottomrule
\end{tabular}
}
\caption{ Assessing the Diversity of Samples From Our Proposed Method}
\label{tab:variance_comparison}
\end{table*}
In this section, we demonstrate that our simpler yet meticulously modified \textit{SimDiff} model effectively leverages diffusion in end-to-end training, accurately capturing distribution shifts while maintaining a balance between sample diversity and accuracy, thereby improving performance through ensembling.

Baseline diffusions illustrate two extremes. TimeDiff \cite{shen2023nonautoregressiveconditionaldiffusionmodels} conditions on a \emph{pre-trained} autoregressive model; this curbs variance but injects bias and limits flexibility. TimeGrad \cite{timegrad} and CSDI \cite{tashiro2021csdi}, in contrast, optimize likelihood only: they generate highly diverse but poorly aligned samples, suffer from training instability and insufficient contexts, and often produce outliers that hurt point forecasts.

SimDiff’s single Transformer denoiser avoids both pitfalls.  A fully end-to-end training process allows the model to harness diffusion’s strengths. By addressing the instability and misaligned distributions observed in TimeGrad and CSDI with tailored design and robust training objectives, our approach ensures that the samples are both diverse and meaningful for ensembling. 

Table \ref{tab:variance_comparison} reports single-shot MSE and ensemble MSE$_E$ (underline indicates improvement); the right-most column lists sample variance averaged over features and horizons. SimDiff attains lower error and controlled variance, confirming that its samples capture temporal structure more faithfully than those of the baselines. Appendix material and Section \ref{sec:method} detail why the one-stage design produces richer samples than pre-training pipelines and how our design mitigates OOD drift, reduces bias, and stabilizes optimization. 
Together, these components let SimDiff deliver the most reliable distribution estimates and the strongest point forecasts after ensembling. In the following sections, we will delve deeper into how these modules contribute to our model's performance.

\subsection{Significance of Training-Specific Normalization Independence}\label{subsec:niab}
\begin{table}[htbp]
\centering
\resizebox{\columnwidth}{!}{
\begin{tabular}{c|cccccccc}
\toprule
\textbf{N.I.} & \textit{NorPool} & \textit{Elec.} & \textit{Traffic} & \textit{ETTm1} & \textit{Weather} & \textit{Wind} & \textit{Exchange} \\
\midrule
\cmark & $\textbf{0.534}$ & $\textbf{0.145}$ & $\textbf{0.383}$ & $\textbf{0.322}$ & $\textbf{0.299}$ & $\textbf{0.880}$ & $\textbf{0.015}$ \\
\xmark & $0.555$ & $0.151$ & $0.389$ & $0.327$ & $0.328$ & $0.891$ & $0.019$ \\
\bottomrule
\end{tabular}
}
\caption{Ablation study showing the essential role of N.I.}
\label{tab:normalization_ablation}
\end{table}

We conduct an ablation study to verify the impact of our proposed N.I. . This technique is designed to mitigate out-of-distribution (OOD) issues by reducing the distributional bias between past observations and future targets.

The results, as shown in Table \ref{tab:normalization_ablation}, indicate that N.I. consistently improves the model’s performance across various datasets, especially for the dataset with severe O.O.D. like \textit{Weather} and \textit{NorPool}, as also can be seen in Figure \ref{fig:niood}.

These results underscore the significance Normalization Independence. During training we “Gaussianise’’ each segment with its \emph{own} statistics, which stabilises optimisation, while a learnable affine layer on the past sequence predicts the future shift / scale at test time. This simple change cuts bias, handles OOD drift, and lowers MSE—ultimately contributing to the success of our ensemble strategy. Also, we provide a preliminary proof in the appendix material.
\subsection{Ensemble Enhances SimDiff as an Effective Point Estimator}
\begin{figure}[ht] 
\centering
\begin{subfigure}[t]{0.31\linewidth} 
    \centering
    \includegraphics[width=\linewidth]{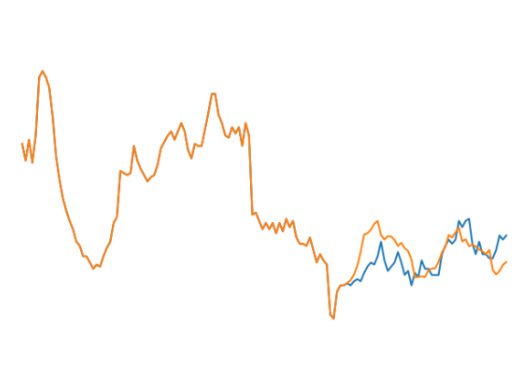}
    \caption{MoM}
    \label{fig:fig1}
\end{subfigure}
\hspace{0.04in}
\begin{subfigure}[t]{0.31\linewidth} 
    \centering
    \includegraphics[width=\linewidth]{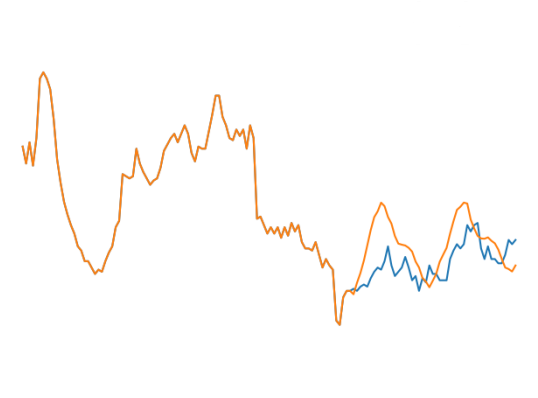}
    \caption{Simple Avg.}
    \label{fig:fig2}
\end{subfigure}
\hspace{0.04in}
\begin{subfigure}[t]{0.31\linewidth} 
    \centering
    \includegraphics[width=\linewidth]{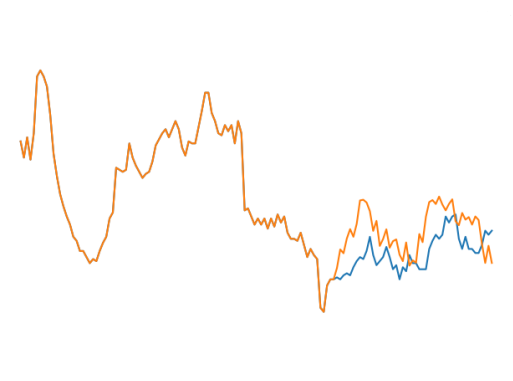}
    \caption{1 Inference}
    \label{fig:fig3}
\end{subfigure}
\caption{Power of MoM}
\label{fig:figmom}
\end{figure}

\begin{table}[h]
\centering
\resizebox{.38\textwidth}{!}{
\begin{tabular}{c|cccc}
\toprule
\textbf{Ensemble}& \textit{ETTh1} & \textit{Weather}& \textit{Wind}& \textit{Caiso} \\
\midrule
\textit{MoM} & $\textbf{0.394}$  &  $\textbf{0.299}$ & $\textbf{0.880}$  & $\textbf{0.106}$  \\
\textit{Avg.} &  $0.398$ & $0.305$  & $0.887$  &  $0.109$ \\
\textit{1 Inf.} & $0.408$  &  $0.317$ & $0.901$  & $0.110$  \\
\bottomrule
\end{tabular}
}
\caption{Impact of Various Ensemble Methods on Boosting Effectiveness}
\end{table}
We ablated three strategies—single-sample inference, simple averaging, and our Median-of-Means ensemble—to gauge their effect on accuracy and stability.
All three ensembles lift SimDiff’s accuracy, but MoM delivers the largest gain.  Simple averaging smooths away high-frequency detail, whereas MoM effectively captures the true distribution of the data, retaining subtle temporal patterns rather than a smoothed trajectory (Figure \ref{fig:figmom}). MoM's robustness to outliers and heavy-tailed noise significantly improves prediction stability, leading to the lowest MSE across all datasets. Theoretically, MoM also offers stronger statistical guarantees, providing tighter concentration bounds in finite-sample regimes, as formally proven in Appendix material.
By replacing the single deterministic pass used in pretrain–conditioned diffusions \cite{shen2023nonautoregressiveconditionaldiffusionmodels,shen2024multiresolution} with MoM ensemble, SimDiff exploits the full probalilistic trace of diffusion while retaining numerical stability.
\subsection{Inference Efficiency}
As shown in Table \ref{tab:ef}, the simple and effective design of our model brings over 90\% improvement in inference speed. \textit{SimDiff} ranks first in single-sample inference time among all evaluated diffusion models, underscoring the efficiency of our single-stage Transformer architecture.

\begin{table}[h]
\centering
\resizebox{\linewidth}{!}{
\begin{tabular}{l| ccccc}
\toprule
& $H=96$ & $H=168$ & $H=192$ & $H=336$ & $H=720$ \\
\midrule
\textbf{SimDiff}  & \textbf{0.22} & \textbf{0.24} & \textbf{0.30} & \textbf{0.33} & \textbf{0.46} \\
\midrule
TimeDiff & 4.73 & 5.21 & 5.84 & 6.91 & 8.40\\
mr-Diff  & 7.02 & 7.41 & 8.10 & 8.95 & 10.92 \\
CSDI & 67.02 & 89.36 & 108.41 & 295.15 & 379.80 \\
TMDM & 111.23 & 152.92 & 187.24 & 252.18 & 483.39 \\
SSSD & 135.92 & 204.25 & 236.32 & 368.29 & 886.56 \\
TimeGrad & 294.85 & 573.05 & 621.70 & 1071.45 & 2312.26 \\
\bottomrule
\end{tabular}
}
\caption{Inference time (in ms) of various time series diffusion models in single inference process with different prediction horizons ($H$) on the \textbf{ETTh1}.}\label{tab:ef}
\end{table}

This efficiency is particularly notable when compared to the next fastest model, TimeDiff. While TimeDiff reduces some computational load by pre-training a regressor, it still relies on a more complex U-Net for its denoising block. In contrast, our streamlined Transformer design avoids this architectural overhead. Consequently, even when \textit{SimDiff} performs multiple inference passes for our MoM ensemble, the total computational time remains highly competitive, striking an balance between accuracy and practical efficiency.

Further analyses, including ablations on model components and parameter sensitivity, can be found in the Appendix.

\section{Related Works}

\subsection{Diffusion Models for Time Series Forecasting} 
Diffusion models, initially developed for generation (DDPM~\cite{NEURIPS2020_4c5bcfec}), are now applied to time series forecasting. Early autoregressive models like TimeGrad~\cite{timegrad} suffered from cumulative errors and slow inference. Non-autoregressive approaches (CSDI~\cite{tashiro2021csdi}, SSSD~\cite{alcaraz2022diffusion}) address these issues but still experience boundary artifacts and computational overhead. Recent methods such as TimeDiff~\cite{shen2023nonautoregressiveconditionaldiffusionmodels} and mr-Diff~\cite{shen2024multiresolution} use conditional and multiresolution strategies to enhance accuracy. However, their reliance on pretrained bases like DLinear~\cite{zeng2022transformers} limits adaptability to complex series and reduces generative flexibility.

\subsection{Transformers and Diffusion Transformers} 
\textbf{Transformers for Time Series Forecasting.} Transformers significantly improve forecasting through long-range dependency modeling, such as Informer~\cite{zhou2021informer}, Autoformer~\cite{wu2021autoformer}, PatchTST~\cite{Yuqietal-2023-PatchTST}, and FEDformer~\cite{22fedformer}.

\noindent \textbf{Diffusion Transformers.} Recent transformer-based diffusion models, such as U-ViT~\cite{bao2022all} and DiT~\cite{peebles2023scalable}, improve scalability and generative performance. However, designed primarily for static vision data, they face challenges capturing temporal dependencies in time series.

\subsection{Other Deep Learning Models}
Other deep learning approaches include basis-expansion methods (FiLM~\cite{zhou2022film}, NBeats~\cite{oreshkin2019n}, DUET~\cite{qiu2025duet},DAG~\cite{qiu2025dag},DBLoss~\cite{qiu2025DBLoss}), hybrid recursive convolutional models (SCINet~\cite{liu2022scinet}), CNN-based approaches (TimesNet~\cite{wu2022timesnet}, DeepGLO~\cite{sen2019think}), and RNN-based models (LSTNet~\cite{lai2018modeling}, DeepAR~\cite{salinas2020deepar}). These models balance interpretability, scalability, and accuracy differently, each with inherent trade-offs.

\section{Conclusion}
In this work, we present \textit{SimDiff}, a simple yet effective diffusion model for time series point forecasting. SimDiff addresses the twin challenges of providing sufficient contextual bias for stability and accuracy and balancing diversity with accuracy, integrating a tailored Transformer within a diffusion framework \emph{without relying on any external pre-trained or jointly trained models}. Its fully end-to-end training achieves state-of-the-art point forecasting and competitive probabilistic performance. Normalization independence and the Median-of-Means estimator further improve robustness to noise and distribution shifts. Owing to its simplicity, SimDiff also delivers faster inference than prior diffusion-based models. We hope these results stimulate further research into diffusion techniques for time series forecasting across diverse domains.

\section*{Acknowledgments}

This work was supported by National Natural Science Foundation of China (NSFC) grants W2441021, 72371172, 72342023, and 71929101. We also thank all authors for their sincere contributions and discussions.

\bibliography{aaai2026}

\newpage
\appendix
\onecolumn
\begin{center}
    {\LARGE \textbf{Appendix Table of Contents}}  
\end{center}
\vspace{12pt}  

\large  
\setlength{\itemsep}{18pt}  

\begin{itemize}
    \item \textbf{Preliminaries - Denoising Diffusion Probabilistic Models} \dotfill \pageref{sec:preli}
    \item \textbf{Training From Scratch Versus Pretrained Methodologies} \dotfill \pageref{sec:pret}
    \item \textbf{Why Do We Normalize Independently} \dotfill \pageref{sec:wdwni}
    \item \textbf{Proof of MoM} \dotfill \pageref{appendix:proof}
    \item \textbf{More Information About Experiments} \dotfill \pageref{app:exps}
    \begin{itemize}
        \setlength{\itemsep}{16pt}  
        \item \textbf{Probabilistic Forecasting Settings} \dotfill \pageref{subsec:pfs}
        \item \textbf{Point Forecasting Settings} \dotfill \pageref{subsec:point}
        \item \textbf{The Critical Role of RoPE in Diffusion Models} \dotfill \pageref{subsec:ropee}
        \item \textbf{Tailored Diffusion-Transformer Forecasting Model} \dotfill \pageref{subsec:tdit}
        \item \textbf{Sensitivity Analysis} \dotfill \pageref{subsec:sae}
        \item \textbf{Inference Times and MoM Settings} \dotfill \pageref{subsec:ifms}
    \end{itemize}
    \item \textbf{Testing MAE for the multivariate setting} \dotfill \pageref{sec:maes}
    \item \textbf{Limitations and Future Directions} \dotfill \pageref{sec:lims}
\end{itemize}

\newpage
\section{Preliminaries - Denoising Diffusion Probabilistic Models}\label{sec:preli}

Denoising Diffusion Probabilistic Models (DDPMs)~\cite{NEURIPS2020_4c5bcfec} are prominent in the field of generative modeling due to their efficacy in sequentially refining data from noise. The forward diffusion process progressively corrupts an initial input $\mathbf{x}_0$ into a Gaussian noise vector through a series of steps. At each step $k$, the state $\mathbf{x}_k$ is derived from its predecessor $\mathbf{x}_{k-1}$ by adding Gaussian noise with variance $\beta_k \in [0, 1]$. This process can be mathematically expressed as:
\begin{equation}
    q(\mathbf{x}_k|\mathbf{x}_{k-1}) = \mathcal{N}(\mathbf{x}_k; \sqrt{1 - \beta_k}\mathbf{x}_{k-1}, \beta_k\mathbf{I}), \quad k = 1, \ldots, K.
\end{equation}
This transformation can also be described in relation to the initial input $\mathbf{x}_0$ as:
\begin{equation}
    q(\mathbf{x}_k|\mathbf{x}_0) = \mathcal{N}(\mathbf{x}_k; \sqrt{\bar{\alpha}_k}\mathbf{x}_0, (1 - \bar{\alpha}_k)\mathbf{I}),
\end{equation}
where $\bar{\alpha}_k = \prod_{s=1}^{k} \alpha_s$ and $\alpha_k = 1 - \beta_k$. Thus, $\mathbf{x}_k$ can be directly computed as:
\begin{equation}
    \mathbf{x}_k = \sqrt{\bar{\alpha}_k}\mathbf{x}_0 + \sqrt{1 - \bar{\alpha}_k}\epsilon,
\end{equation}
where $\epsilon$ is drawn from $\mathcal{N}(0, \mathbf{I})$. This formulation facilitates the recovery of $\mathbf{x}_0$ from $\mathbf{x}_k$ during the denoising process.

The backward denoising process is framed as a Markov chain. At the $k$-th denoising step, $\mathbf{x}_{k-1}$ is sampled from:
\begin{equation}
    p_{\theta}(\mathbf{x}_{k-1}|\mathbf{x}_k) = \mathcal{N}(\mathbf{x}_{k-1}; \mu_{\theta}(\mathbf{x}_k, k), \Sigma_{\theta}(\mathbf{x}_k, k)).
\end{equation}
Typically, the variance $\Sigma_{\theta}(\mathbf{x}_k, k)$ is fixed to $\sigma_k^2 \mathbf{I}$, while the mean $\mu_{\theta}(\mathbf{x}_k, k)$ is parameterized by a neural network. This process can be posed as either a noise estimation or data prediction task~\cite{benny2022dynamic}.

For noise estimation, a neural network $\epsilon_{\theta}$ predicts the noise component in $\mathbf{x}_k$, and $\mu_{\theta}(\mathbf{x}_k, k)$ is computed as:
\begin{equation}
    \mu_{\theta}(\mathbf{x}_k, k) = \frac{1}{\sqrt{\alpha_k}}\mathbf{x}_k - \frac{1 - \alpha_k}{\sqrt{1 - \bar{\alpha}_k}}\epsilon_{\theta}(\mathbf{x}_k, k).
\end{equation}
The parameters $\theta$ are optimized by minimizing the following loss function:
\begin{equation}
    L_{\epsilon} = \mathbb{E}_{k, \mathbf{x}_0, \epsilon} \left[ \|\epsilon - \epsilon_{\theta}(\mathbf{x}_k, k)\|^2 \right].
\end{equation}
Alternatively, for data prediction, a denoising network $x_{\theta}$ estimates the clean data $\mathbf{x}_0$ from $\mathbf{x}_k$. The mean $\mu_{\theta}(\mathbf{x}_k, k)$ is given by:
\begin{equation}
    \mu_{\theta}(\mathbf{x}_k, k) = \frac{\sqrt{\alpha_k}(1 - \bar{\alpha}_{k-1})}{1 - \bar{\alpha}_k} \mathbf{x}_k + \frac{\sqrt{\bar{\alpha}_{k-1}\beta_k}}{1 - \bar{\alpha}_k} x_{\theta}(\mathbf{x}_k, k).
\end{equation}
The corresponding loss function is:
\begin{equation}
    L_x = \mathbb{E}_{\mathbf{x}_0, \epsilon, k} \left[ \|\mathbf{x}_0 - x_{\theta}(\mathbf{x}_k, k)\|^2 \right].
\end{equation}

\section{Training From Scratch Versus Pretrained Methodologies}\label{sec:pret}

Pre-trained conditioning, while providing a regression-based distributional foundation, limits the generative flexibility of diffusion models, restricting their ability to fully capture the complexity of time series data. This section formalizes the effect of conditioning on variance, highlighting the impact of using a pre-trained model versus training from scratch.

Consider a diffusion model \( p_{\theta}(x_{t} \mid x_{t-1}, c) \), where \( x_{t} \) is the state at time \( t \), \( x_{t-1} \) is the state at time \( t-1 \), and \( c \) is conditioning information provided by a pre-trained model.

\textit{\textbf{Proposition 1.}} \textit{Conditioning on the output \( c \) of a pre-trained model reduces the variance of the predicted state \( x_t \) given \( x_{t-1} \). Formally,
\begin{align}
\text{Var}(x_t \mid x_{t-1}, c) \leq \text{Var}(x_t \mid x_{t-1}).
\end{align}
}

\textit{Proof:}
When training the diffusion model from scratch, we aim to model the true distribution of \( x_t \) given \( x_{t-1} \) without additional conditioning. This involves learning both the mean and variance parameters directly from the data:
\begin{align}
p_\theta(x_t \mid x_{t-1}) \sim \mathcal{N}(\mu_t(x_{t-1}), \sigma_t^2(x_{t-1})).
\end{align}

In contrast, when using a pre-trained model's output as conditioning information \( c \), the diffusion model becomes:
\begin{align}
p_\theta(x_t \mid x_{t-1}, c) \sim \mathcal{N}(\mu_t(x_{t-1}, c), \sigma_t^2(x_{t-1}, c)).
\end{align}

According to the law of total variance, we have:
\begin{align}
\text{Var}(x_t \mid x_{t-1}) &= \mathbb{E}_{c \sim p_\phi(c \mid x_{t-1})} \left[ \text{Var}(x_t \mid x_{t-1}, c) \right]  \notag\\
&+\text{Var}_{c \sim p_\phi(c \mid x_{t-1})} \left[ \mathbb{E}[x_t \mid x_{t-1}, c] \right],
\end{align}
where \( p_\phi(c \mid x_{t-1}) \) is the distribution implied by the pre-trained model.

Since variance is a non-negative quantity, the second term \( \text{Var}_{c} \left[ \mathbb{E}[x_t \mid x_{t-1}, c] \right] \geq 0 \). This implies:
\begin{align}
\mathbb{E}_{c} \left[ \text{Var}(x_t \mid x_{t-1}, c) \right] \leq \text{Var}(x_t \mid x_{t-1}).
\end{align}

Moreover, for any specific \( c \), it holds that:
\begin{align}
\text{Var}(x_t \mid x_{t-1}, c) \leq \text{Var}(x_t \mid x_{t-1}).
\end{align}

This inequality is justified because conditioning on additional information \( c \) cannot increase the variance of \( x_t \) given \( x_{t-1} \); it either reduces it or leaves it unchanged.

Therefore, using a pre-trained model's output \( c \) as conditioning reduces (or at least does not increase) the variance of the diffusion model's predictions compared to training from scratch without such conditioning.

\section{Why Do We Normalize Independently?}\label{sec:wdwni}

Here's a brief proof of why we normalize the series independently. Basically, we follow the notation from section \ref{subsec:NI}.

\textit{\textbf{Proposition 2.}} \textit{We should not normalize the past and future time series using the same statistics due to distribution drift between them during the training period.}

\textit{Proof:}
Consider the traditional normalization approach where both $\mathbf{X}$ and $\mathbf{Y}$ are normalized using $\mu_{\mathbf{X}}$ and $\sigma_{\mathbf{X}}$ computed from $\mathbf{X}$:
\begin{small}
\begin{align}
\begin{aligned}
\mathbf{X}_{\text{norm}} &= \frac{\mathbf{X} - \mu_{\mathbf{X}}}{\sigma_{\mathbf{X}}}, &
\mathbf{Y}_{\text{norm}} &= \frac{\mathbf{Y} - \mu_{\mathbf{X}}}{\sigma_{\mathbf{X}}}.
    \end{aligned}
\end{align}
\end{small}

Any distribution shift between $\mathbf{X}$ and $\mathbf{Y}$ introduces a bias in the normalized future data:
\begin{small}
\begin{align}
    \mathbf{Y}_{\text{norm}} &= \frac{\mathbf{Y} - \mu_{\mathbf{X}}}{\sigma_{\mathbf{X}}} = \frac{ (\mathbf{Y} - \mu_{\mathbf{Y}}) + (\mu_{\mathbf{Y}} - \mu_{\mathbf{X}}) }{ \sigma_{\mathbf{X}} } \notag \\
    &= \frac{ \mathbf{Y} - \mu_{\mathbf{Y}} }{ \sigma_{\mathbf{X}} } 
       + \frac{ \mu_{\mathbf{Y}} - \mu_{\mathbf{X}} }{ \sigma_{\mathbf{X}} } \notag \\
    &= \frac{ \sigma_{\mathbf{Y}} \cdot \mathbf{Y}_{\text{real\_norm}} }{ \sigma_{\mathbf{X}} } 
       + \frac{ \mu_{\mathbf{Y}} - \mu_{\mathbf{X}} }{ \sigma_{\mathbf{X}} }.
\end{align}
\end{small}

Here, $\mu_{\mathbf{Y}} - \mu_{\mathbf{X}}$ represents the distribution shift, and $\mathbf{Y}_{\text{real\_norm}}$ is $\mathbf{Y}$ normalized by its own statistics. Under traditional normalization, the model minimizes the loss:
\begin{small}
\begin{align}
    \mathcal{L}_{\text{trad}}(\theta) 
    &= \mathbb{L}\left( \left\| \mathbf{Y}_{\text{norm}} - f_{\theta}(\mathbf{X}_{\text{norm}}) \right\| \right).
\end{align}
\end{small}

where $f_{\theta}$ is the model parameterized by $\theta$. After denormalization, the predicted future sequence is:
\begin{small}
\begin{align}
    \hat{\mathbf{Y}}_{\text{trad}} &= \sigma_{\mathbf{X}} f_{\theta}(\mathbf{X}_{\text{norm}}) + \mu_{\mathbf{X}}.
\end{align}
\end{small}

The expected prediction error includes the bias term due to distribution drift:
\begin{small}
\begin{align}
    \mathbb{E}\left[ \left\| \hat{\mathbf{Y}}_{\text{trad}} - \mathbf{Y} \right\|^2 \right] &= \mathbb{E}\left[ \left\| \sigma_{\mathbf{X}} f_{\theta}(\mathbf{X}_{\text{norm}}) + \mu_{\mathbf{X}} - \mathbf{Y} \right\|^2 \right] \notag \\
    &= \mathbb{E}\left[ \left\| \sigma_{\mathbf{X}} \left( f_{\theta}(\mathbf{X}_{\text{norm}}) - \mathbf{Y}_{\text{norm}} \right) + (\mu_{\mathbf{X}} - \mu_{\mathbf{Y}}) \right\|^2 \right].
\end{align}
\end{small}

This error includes the bias term $(\mu_{\mathbf{X}} - \mu_{\mathbf{Y}})$, which can be significant when there is distribution drift.

Under Normalization Independence, $\mathbf{X}$ and $\mathbf{Y}$ are normalized independently, as shown in Equation (20)(21).

The model then minimizes the loss:
\begin{small}
\begin{align}
    \mathcal{L}_{\text{NI}}(\theta, \boldsymbol{\gamma}, \boldsymbol{\beta}) &= \mathbb{L}\left( \left\| \mathbf{Y}_{\text{norm}} - f_{\theta, \boldsymbol{\gamma}, \boldsymbol{\beta}}(\mathbf{X}_{\text{norm}}) \right\| \right),
\end{align}
\end{small}

After denormalization, the predicted future sequence can be regarded as:
\begin{small} \begin{align} 
\hat{\mathbf{Y}}_{\text{NI}, t} = \sigma_{\mathbf{Y}, t} f_{\theta, \boldsymbol{\gamma}, \boldsymbol{\beta}}(\mathbf{X}_{\text{norm}})_t + \mu_{\mathbf{Y}, t}.
\end{align} \end{small}

So the expected prediction error is:
\begin{small} \begin{align} \mathbb{E}\left[ \left\| \hat{\mathbf{Y}}_{\text{NI}} - \mathbf{Y} \right\|^2 \right] = \mathbb{E}\left[ \left\| \sigma_{\mathbf{Y}, t} \left( f_{\theta, \boldsymbol{\gamma}, \boldsymbol{\beta}}(\mathbf{X}_{\text{norm}})_t - \mathbf{Y}_{\text{norm}, t} \right) \right\|^2 \right]. \end{align} \end{small}

Since $\mathbf{Y}_{\text{norm}, t}$ is normalized using its own statistics, the bias introduced by distribution drift is eliminated.

The error difference between the two methods is:
\begin{small}
\begin{align}
    \Delta \mathbb{E} &= \mathbb{E}\left[ \left\| \hat{\mathbf{Y}}_{\text{trad}} - \mathbf{Y} \right\|^2 \right] - \mathbb{E}\left[ \left\| \hat{\mathbf{Y}}_{\text{NI}} - \mathbf{Y} \right\|^2 \right] \notag \\
     &= \mathbb{E}\left[ \left\| \sigma_{\mathbf{X}} \left( f_{\theta}(\mathbf{X}_{\text{norm}}) - \mathbf{Y}_{\text{norm}} \right) + (\mu_{\mathbf{X}} - \mu_{\mathbf{Y}}) \right\|^2 \right] \notag \\
        &\quad - \mathbb{E}\left[ \left\| \boldsymbol{\sigma}_{\mathbf{Y}, t}  \left( f_{\theta, \boldsymbol{\gamma}, \boldsymbol{\beta}}(\mathbf{X}_{\text{norm}})_t - \mathbf{Y}_{\text{norm}, t} \right) \right\|^2 \right].
\end{align}
\end{small}

Thus, Normalization Independence reduces the systematic error by eliminating the bias term due to distribution drift; Also, the trainable parameters $\gamma$,$\beta$ help reduce the prediction error, resulting in a lower expected prediction error, allowing the model to better adapt to changes in data distribution and improving the accuracy and robustness of future predictions.

\section{Proof of MoM}\label{appendix:proof}

Continuing with the notation in section 3.4, here we provide a detailed proof. To summarize, the event \(|\hat{\mu}_{MoM} - \mu_0| > \epsilon\) implies that at least half of the \(\hat{\mu}_k\) deviates from \(\mu_0\) by more than \(\epsilon\). Formally,
\[
\left\{|\hat{\mu}_{MoM} - \mu_0| > \epsilon \right\} \subset \left\{ \sum_{k=1}^{K} I(|\hat{\mu}_k - \mu_0| > \epsilon) \geq \frac{K}{2} \right\}
\]
where \(\hat{\mu}_{MoM}\) is the Median of Means estimator, \(\mu_0\) is the true mean, \(\hat{\mu}_k\) are the mean estimates of the \(K\) subsamples, \(\epsilon\) is the deviation threshold, and \(I(\cdot)\) is the indicator function which equals 1 if the condition inside is true and 0 otherwise.

Let \(Z_k = I(|\hat{\mu}_k - \mu_0| > \epsilon)\) and \(p_{\epsilon, B} = E(Z_k) = P(|\hat{\mu}_k - \mu_0| > \epsilon)\), where \(E(Z_k)\) is the expectation of \(Z_k\) and \(P(|\hat{\mu}_k - \mu_0| > \epsilon)\) is the probability that \(|\hat{\mu}_k - \mu_0|\) exceeds \(\epsilon\). Therefore,
\[
P(|\hat{\mu}_{MoM} - \mu_0| > \epsilon) \leq P \left( \sum_{k=1}^{K} Z_k \geq \frac{K}{2} \right)
\]

Applying Hoeffding's inequality, which provides an upper bound on the probability that the sum of bounded independent random variables deviates from its expected value, we obtain:
\[
P \left( \frac{1}{K} \sum_{k=1}^{K} (Z_k - E(Z_k)) \geq t \right) \leq e^{-2Kt^2}
\]
Thus,
\[
P \left( |\hat{\mu}_{MoM} - \mu_0| > \epsilon \right) \leq P \left( \frac{1}{K} \sum_{k=1}^{K} (Z_k - E(Z_k)) \geq \frac{1}{2} - p_{\epsilon, B} \right) \leq e^{-2K \left( \frac{1}{2} - p_{\epsilon, B} \right)^2}
\]

Given that the variance \(\sigma^2 = \text{Var}(X_1) < \infty\), by Chebyshev's inequality, which states that for any random variable with finite mean and variance, the probability that the variable deviates from its mean by more than a specified number of standard deviations is bounded, we have:
\[
p_{\epsilon, B} = P(|\hat{\mu}_k - \mu_0| > \epsilon) \leq \frac{\sigma^2}{B \epsilon^2} = \frac{K \sigma^2}{n \epsilon^2}
\]
Therefore, the above inequality simplifies to:
\[
P(|\hat{\mu}_{MoM} - \mu_0| > \epsilon) \leq e^{-2K \left( \frac{1}{2} - \frac{K \sigma^2}{n \epsilon^2} \right)^2}
\]

\section{More Information About Experiments}\label{app:exps}

\subsection{Probabilistic Forecasting Settings}\label{subsec:pfs}

\noindent \textbf{Datasets:} We evaluate the SimDiff on 4 real-world benchmarks, which are the most popular probabilistic forecasting applications: 1)  Electricity (370 dimensions) of power domain. 2) Traffic (963 dimensions) and 3) Taxi (1214 dimensions) of traffic domain, 4)Wikipedia (2000 dimensions) of click-rate domain. 

\noindent \textbf{Baselines:} In the multivariate probabilistic forecasting task, we include several highly praised baseline methods:  GP-Copula\cite{salinas2019high}, LSRP and LSTM-MAF\cite{rasul2020multivariate}. And for the diffusion based models, including TimeGrad\cite{timegrad}, CSDI\cite{tashiro2021csdi}, SSSD\cite{alcaraz2022diffusion},LDT\cite{feng2024latent},D3VAE\cite{ligenerative}, TMDM\cite{li2024transformermodulated}, MG-TSD\cite{fan2024mgtsdmultigranularitytimeseries} and TSDiff\cite{kollovieh2023predictrefinesynthesizeselfguiding}. As for the forecasting setting of datasets, we basically follow the previous work LDT\cite{feng2024latent} and CSDI\cite{tashiro2021csdi}.

\subsection{Point Forecasting Settings}\label{subsec:point}

The following sections detail our experimental setup, including datasets, baseline models, and configurations used to evaluate SimDiff's performance across various time series forecasting tasks.

\noindent \textbf{Datasets:} Our experiments were conducted on nine real-world time series datasets~\cite{zhou2021informer,wu2021autoformer,fan2022depts,lai2018modeling,ligenerative} as presented in Table \ref{tab:dataset}.

\begin{table*}[bht]
  \caption{\textit{Detailed Dataset Descriptions}. Dim denotes the variate number of each dataset. Dataset Size denotes the total number of time points in (Train, Validation, Test) split respectively. History Length denotes the historical time points as conditions. Prediction Length denotes the future time points to be predicted. Frequency denotes the sampling interval of time points.}
  \label{tab:dataset}
  \vskip 0.05in
  \centering
  \begin{small}
  \setlength{\tabcolsep}{6.5pt}
  \resizebox{\textwidth}{!}{
  \begin{tabular}{l|c|c|c|c|c|l}
    \toprule
    \textbf{Dataset} & \textbf{Dim} & \textbf{History Length} & \textbf{Prediction Length} & \textbf{Dataset Size} & \textbf{Frequency} & \textbf{Information} \\
    \midrule
    \textit{ETTh1} & 7 & 336 & 1 week (168) & (8137, 2713, 2713) & Hourly & Electricity\\
    \midrule
    \textit{ETTm1} & 7 & 1440 & 2 days (192) & (32929, 11329, 11329) & 15 mins & Electricity\\
    \midrule
    \textit{Exchange} & 8 & 96 & 2 weeks (14) & (5202, 747, 1504) & Daily & Economy \\
    \midrule
    \textit{Weather} & 21 & 1440 & 1 week (672) & (34776, 4599, 9868) & 10 mins & Weather\\
    \midrule
    \textit{Electricity} & 321 & 720 & 1 week (168) & (17525, 2465, 5093) & Hourly & Electricity \\
    \midrule
    \textit{Traffic} & 862 & 1440 & 1 week (168) & (10673, 1589, 3341) & Hourly & Transportation \\
    \midrule
    \textit{Wind} & 7 & 1440 & 2 days (192) & (37307, 4677, 4676) & 15 mins & Wind Power \\
    \midrule
    \textit{NorPool} & 18 & 1440 & 1 month (720) & (61447, 3559, 1525) & Hourly & Energy \\
    \midrule
    \textit{Caiso} & 10 & 1440 & 1 month (720) & (58334, 6781, 5760) & Hourly & Energy \\
    \bottomrule
  \end{tabular}
  }
  \end{small}  
  \vspace{-4pt}
\end{table*}

We primarily follow the settings from TimeDiff~\cite{shen2023nonautoregressiveconditionaldiffusionmodels} and mr-Diff~\cite{shen2024multiresolution}. These methods, which are competitive state-of-the-art point estimators of diffusion models, utilize a pretrained autoregressive initialization model. This setting allows the model to select the history length from {96, 192, 336, 720, 1440} using the validation set, thus preventing the underestimation of baselines and leveraging the different strengths of each model to consistently choose the best results, thus creating strong baselines.
Moreover, for clarity and reproducibility, we provide the actual window settings and dataset splits used in our experiments in this section(Table \ref{tab:dataset}).

\noindent \textbf{Baselines and Experimental Settings.} We evaluated a comprehensive range of models across different categories, employing the following as baselines:
(i) \textit{Diffusion Models for Time Series:} mr-Diff~\cite{shen2024multiresolution},TimeDiff~\cite{shen2023nonautoregressiveconditionaldiffusionmodels}, TimeGrad~\cite{timegrad}, CSDI~\cite{tashiro2021csdi}, SSSD~\cite{alcaraz2022diffusion};
(ii) \textit{Generative Models for Time Series:} D3VAE~\cite{ligenerative}, CPF~\cite{rangapuram2023coherent}, PSA-GAN~\cite{jeha2022psa};
(iii) \textit{Prediction Models Based on Basis Expansion:} NHits~\cite{challu2023nhits}, FiLM~\cite{zhou2022film}, Depts~\cite{fan2022depts}, NBeats~\cite{oreshkin2019n};
(iv) \textit{Time Series Transformers:} Scaleformer~\cite{shabani2022scaleformer}, PatchTST~\cite{Yuqietal-2023-PatchTST}, Fedformer~\cite{22fedformer}, Autoformer~\cite{wu2021autoformer}, Pyraformer~\cite{liu2021pyraformer}, Informer~\cite{zhou2021informer}, Transformer~\cite{vaswani2017attention};
(v) \textit{Other Competitive Baselines:} SCINet~\cite{liu2022scinet}, NLinear~\cite{zeng2022transformers}, DLinear~\cite{zeng2022transformers}, and LSTMa~\cite{bahdanau2015neural}, an attention-based LSTM~\cite{hochreiter1997long}. For all models metioned above, in order to avoid under-estimating the baselines and take advantage of the different designs of each model, we follow the setting from TimeDiff~\cite{shen2023nonautoregressiveconditionaldiffusionmodels} and mr-Diff~\cite{shen2024multiresolution}, where we allow the model to select the history length from \{96, 192, 336, 720, 1440\}. 


\noindent \textbf{Implementation Details.} We train our proposed model using the Adam optimizer~\cite{kingma2014adam} with a learning rate of $10^{-3}$.  We employ early stopping with a maximum of 100 epochs. Training involves $K = 100$ diffusion steps, using a cosine noise schedule~\cite{nichol2021improved} with $\beta_t$ in [0, 0.999] and an offset $s = 5$. To speed up diffusion model inference, efficient learning-free samplers like DDIM~\cite{song2021denoising}, Analytic-DPM~\cite{bao2022analyticdpm}, and DPM-Solver~\cite{lu2022dpmsolver} have been developed. In this work, we use DPM-Solver, which reduces the number of denoising steps with minimal impact on performance. Empirically, the number of denoising steps during inference can be reduced to fewer than 5. All experiments are run on one or two Nvidia RTX A100 GPU.

\subsection{The Critical Role of RoPE in Diffusion Models}\label{subsec:ropee}
\begin{figure*}[htbp]
    \centering
    \begin{subfigure}{0.49\textwidth}
        \centering
        \includegraphics[width=\linewidth]{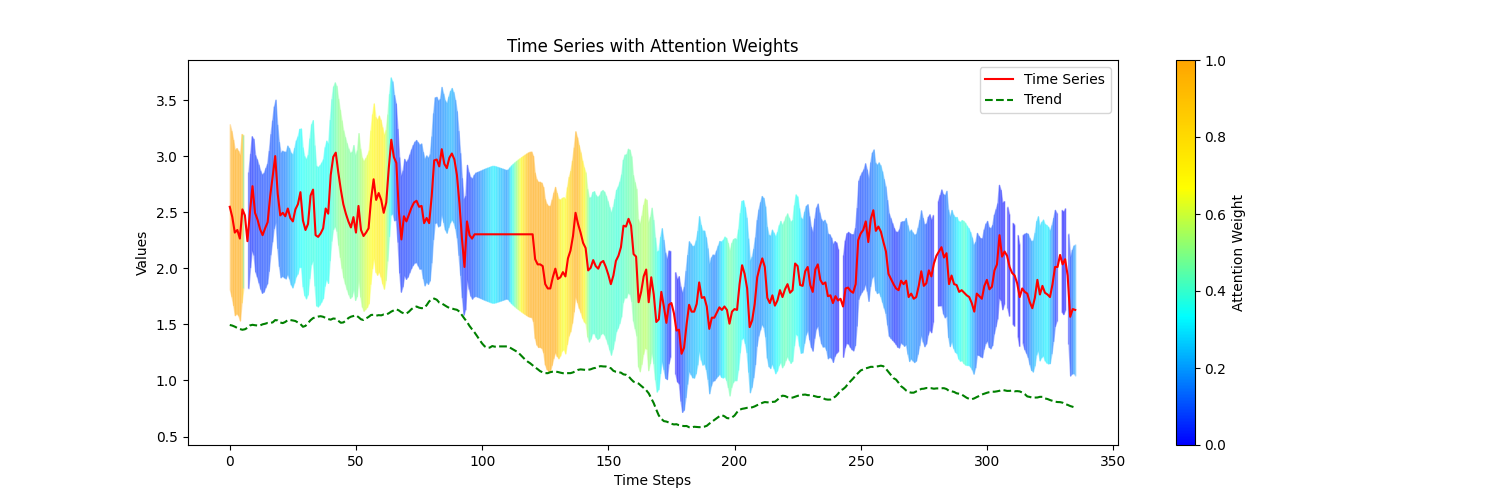} 
        \caption{Attention Weights with RoPE}
    \end{subfigure}
    \begin{subfigure}{0.49\textwidth}
        \centering
        \includegraphics[width=\linewidth]{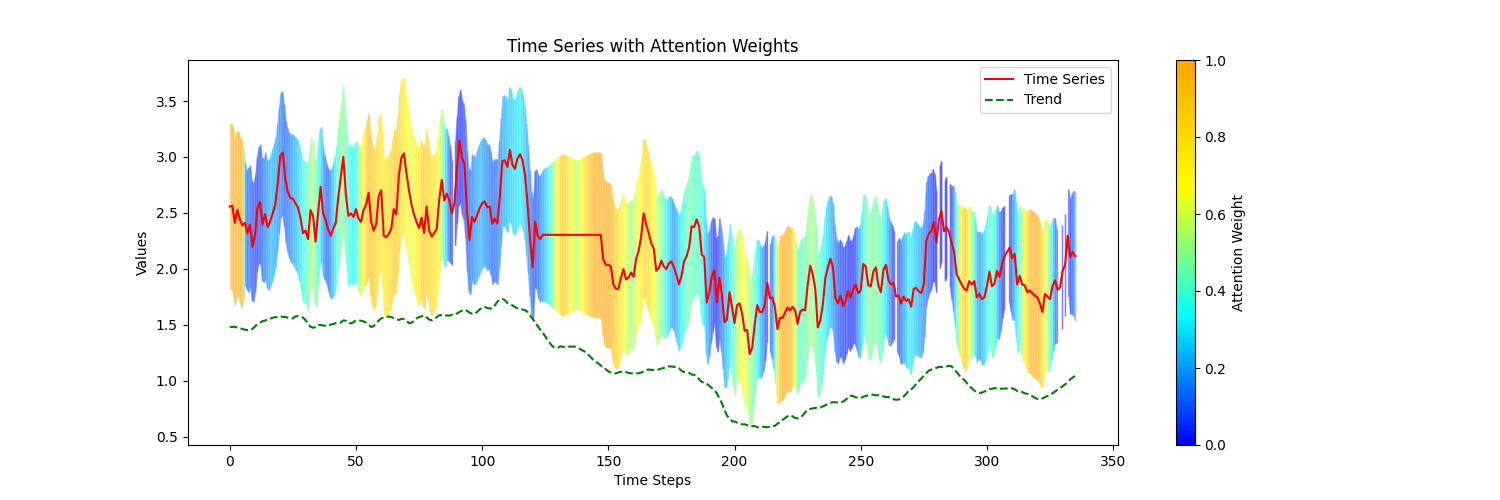} 
        \caption{Attention Weights without RoPE}
    \end{subfigure}
    \vskip -0.1in
    \caption{Enhancing Attention on Distribution Drift with RoPE}\label{fig:aro}
    \vskip -0.1in
\end{figure*}
In this section, We want to emphasize the importance of RoPE in long-term time series forecasting. As shown in Table \ref{tab:rope_ablation} and Figure \ref{fig:aro}, we conducted an ablation experiment to evaluate the impact of introducing RoPE in our \textit{SimDiff} model.
\begin{table}[h]
\centering
\vskip -0.15in
\caption{Ablation Experiment for Introducing the RoPE}
\resizebox{.4\columnwidth}{!}{
\begin{tabular}{c|cccc}
\toprule
\textbf{RoPE}& \textit{ETTh1} & \textit{Weather}& \textit{NorPool}& \textit{ETTm1} \\
\midrule
\cmark & $\textbf{0.394}$  &  $\textbf{0.299}$ & $\textbf{0.534}$  & $\textbf{0.322}$  \\
\xmark &  $0.401$ & $0.310$  & $0.582$  &  $0.328$ \\
\bottomrule
\end{tabular}
}
\label{tab:rope_ablation}
\vskip -0.1in
\end{table}
RoPE’s robust encoding of positional information ensures that the model maintains temporal context, resulting in more precise and reliable forecasts. This encoding enables the transformer backbone to focus attention on patterns associated with distribution drift, learning these conditions more effectively rather than randomly dispersing attention on normal or overly similar patterns, as shown in Figure \ref{fig:aro}. By incorporating RoPE, the \textit{SimDiff} model effectively captures temporal dependencies and relevance in time series data. The consistent performance improvements across multiple datasets highlight the importance of positional information, validating the significance of using RoPE in enhancing long-term forecasting capabilities.

\subsection{Tailored Diffusion-Transformer Forecasting Model}\label{subsec:tdit}
\begin{figure*}[htbp]
    \centering
    \begin{subfigure}{0.3\textwidth}
        \centering
        \includegraphics[width=\linewidth]{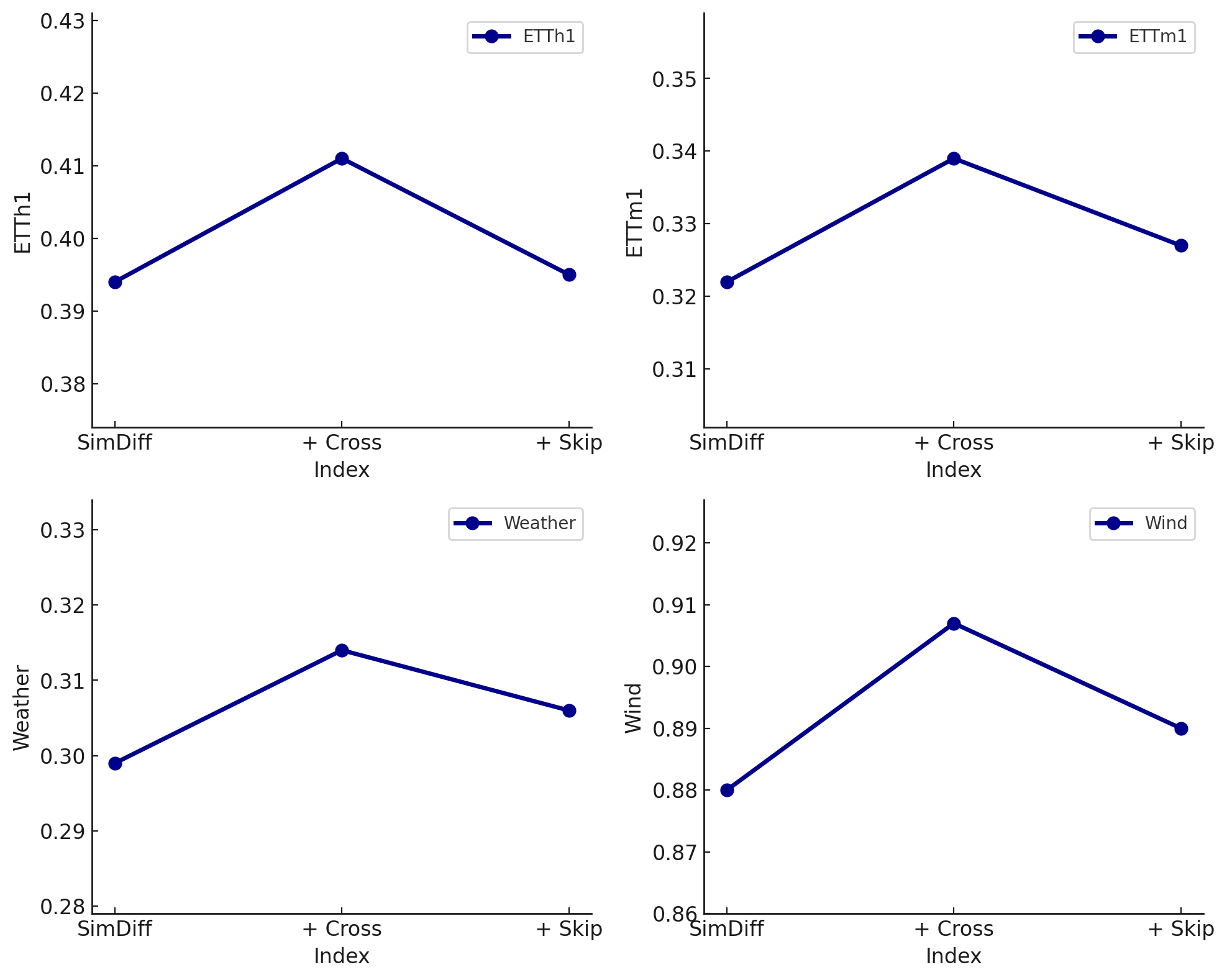} 
        \caption{Transformer Structure}
        \label{fig:stru}
    \end{subfigure}
    \begin{subfigure}{0.3\textwidth}
        \centering
        \includegraphics[width=\linewidth]{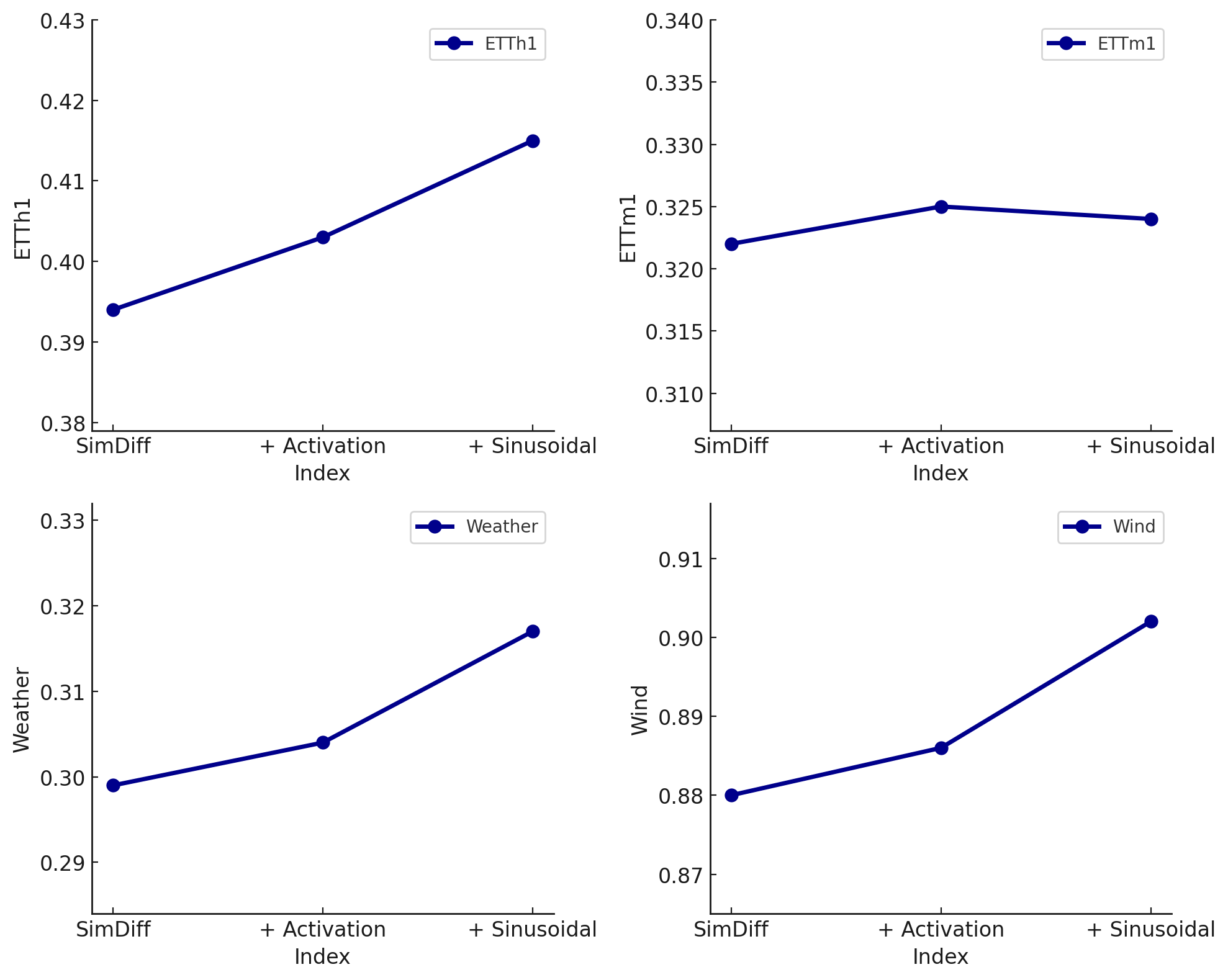} 
        \caption{Time Token Embedding}
        \label{fig:time}
    \end{subfigure}
    \begin{subfigure}{0.3\textwidth}
        \centering
        \includegraphics[width=\linewidth]{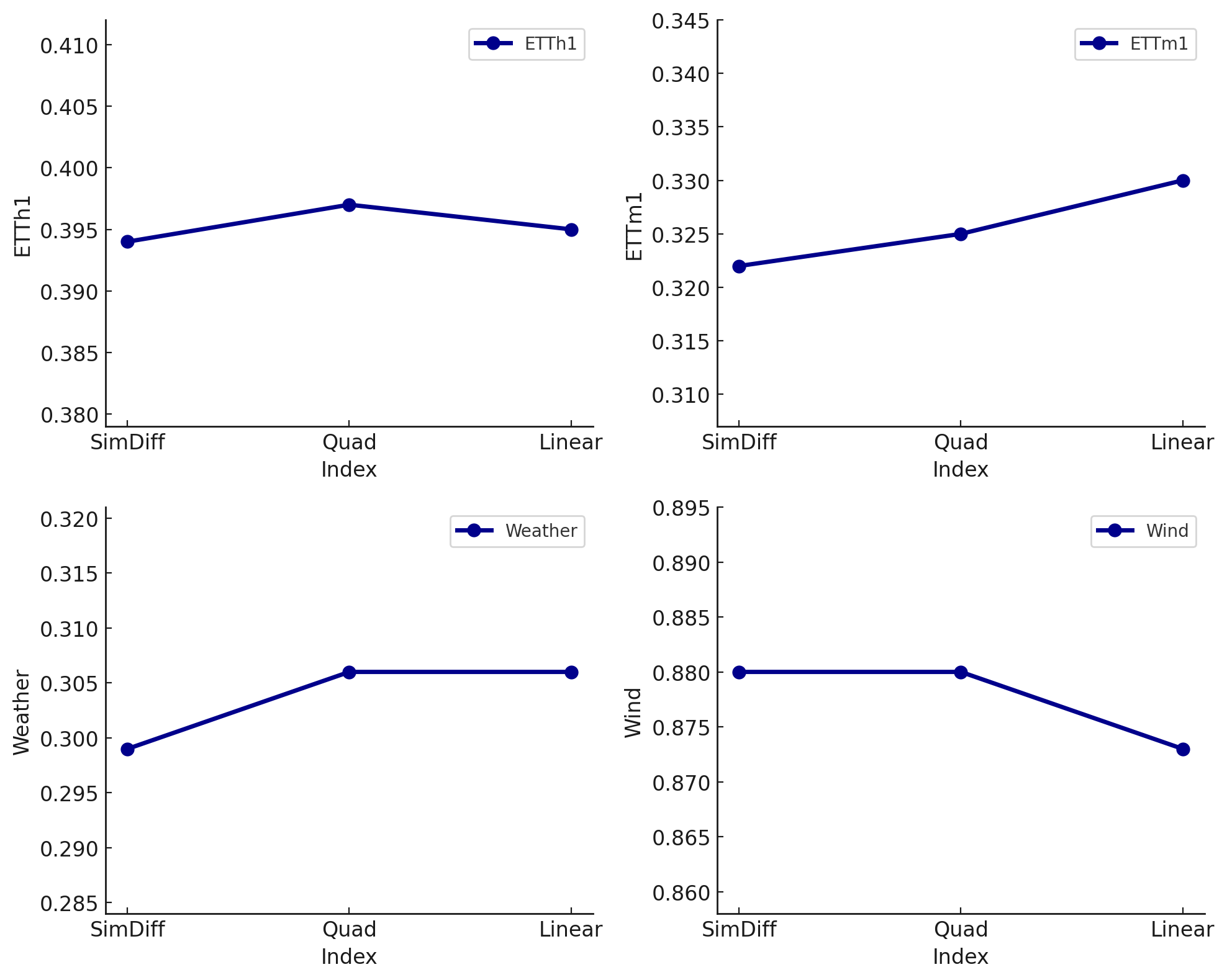} 
        \caption{Noise Schedule Choice}
        \label{fig:sch}
    \end{subfigure}
    \caption{Tailored Diffusion-Transformer Forecasting Model}
    \label{fig:tail}
\end{figure*}

We conducted extensive experiments to design our model, and some of the results are shown in the figure \ref{fig:tail}.

\noindent \textbf{a) Cross-Channel Attention and Skip Connections} As shown in Figure \ref{fig:stru}, simply adding cross-attention and skip connections introduce noisy, unrelated information across channels and time steps, complicating optimization and destabilizing convergence.This interference disrupts the diffusion model’s accurate learning of temporal distributions, making both cross-attention and skip connections suboptimal for our diffusion model.

\noindent \textbf{b) Embedding for Diffusion Timesteps} As illustrated in Figure \ref{fig:time}, the embedding method for diffusion timesteps significantly impacts model performance. We found that a straightforward linear embedding achieved the best results, while adding activation functions or using sinusoidal embedding was less effective. Hence, we opted for linear embedding to maintain simplicity and stability.

\noindent \textbf{c) Noise Scheduling Strategy} As depicted in Figure \ref{fig:sch}, the cosine noise schedule achieves optimal performance by controlling the noise intensity across early and late stages, aligning better with training dynamics. This approach also smoothly distributes noise, preventing excessive concentration at specific timesteps. In contrast, other schedules, such as quadratic and linear, showed inconsistent results across datasets. Thus, we selected cosine scheduling as our default noise strategy.

These tailored design choices enable our diffusion-transformer model to improve forecasting accuracy and adaptability by effectively managing noise and unwanted unstable enhancements.

\subsection{Sensitivity Analysis}\label{subsec:sae}

\begin{figure}[htbp]
    \centering
    \includegraphics[width=0.6\linewidth]{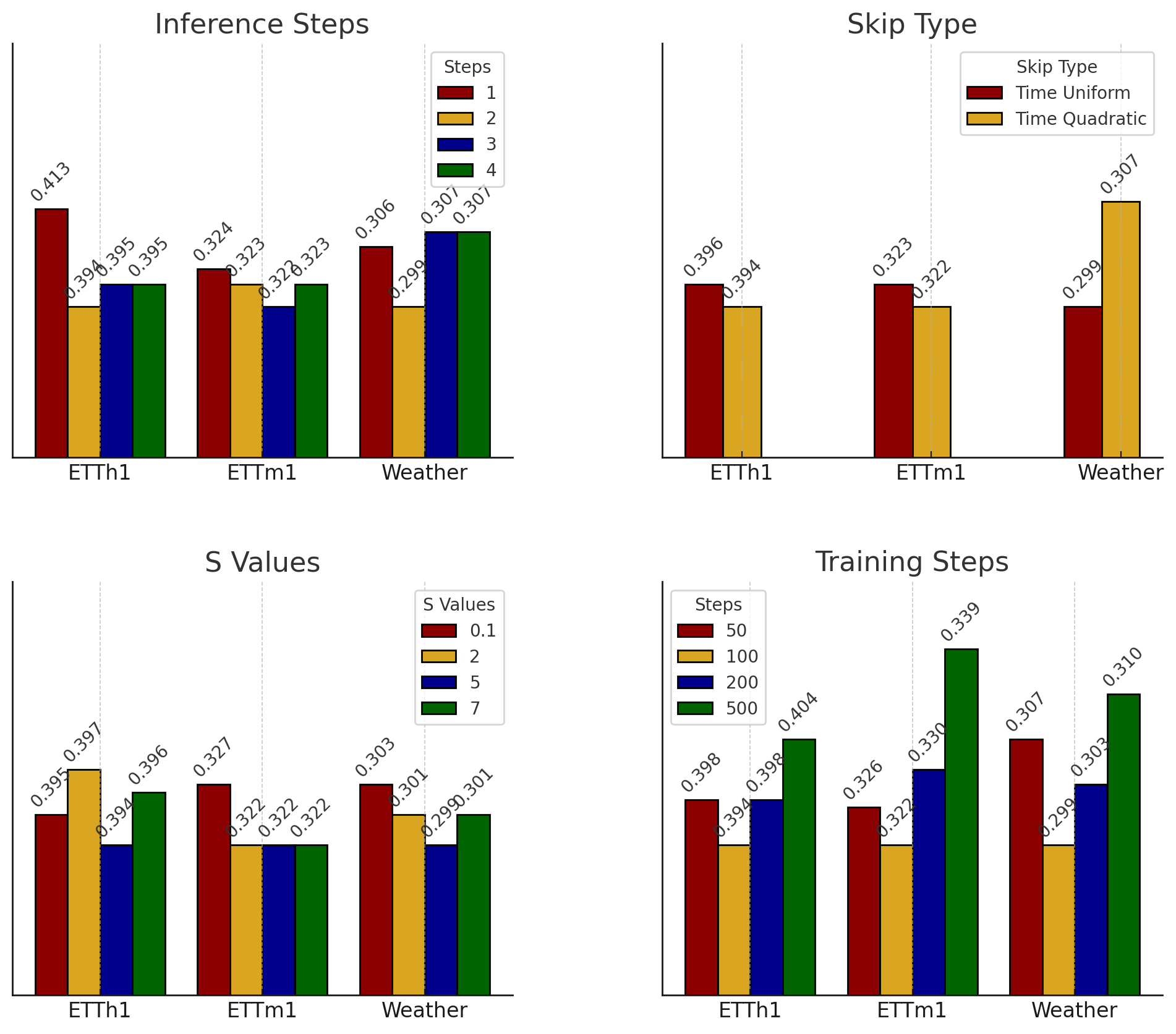} 
    \caption{Sensitivity Analysis}\label{fig:sens}
\end{figure}

In this section, we conduct more parameters sensitivity experiments and model efficiency analysis, as shown in Figure \ref{fig:sens} and Table \ref{tab:ef}.

\noindent \textbf{a) Inference Steps:} Our model’s robustness and strong performance allow for fewer inference steps (typically 2 or 3; 1 step is insufficient), which also generates a more diverse probability distribution with significantly reduced inference time.

\noindent \textbf{b) Skip Type:} For accelerated sampling, we follow the literature's settings except for skip types. Experience from Computer Vision suggests that Time Uniform skips suit high-resolution images, while Time Quadratic works better for low-resolution. This also applies in time series: as shown in Figure \ref{fig:sens}, ETTh1 and ETTm1 (both with 7 variables) favor Time Quadratic, whereas Weather (21 variables) performs best with Time Uniform.

\noindent \textbf{c) Cosine Beta Schedule:} In the cosine noise schedule, the bias parameter $S$ controls noise progression during training, with larger $S$ values introducing weaker initial noise that intensifies gradually. Results show that this "weak first, strong later" noise pattern is more effective for training time series models.

\noindent \textbf{d) Training Diffusion Steps:} The number of diffusion steps during training also significantly impacts performance. Empirically, 100 steps provide optimal results.

\subsection{Inference Times and MoM Settings}\label{subsec:ifms}

Regarding the number of inference runs, the difference between a single inference and multiple inferences is substantial. However, once the number of inference runs reaches approximately 30, the performance improvement from further increasing the number of runs becomes trivial. For numerical convenience, we generally set the number of inference runs to 100.

For the MoM settings, we find that using more than two groups already provides significant improvements. In our experiments, we typically use three or five groups, as increasing the number of groups beyond this does not yield additional benefits. Similarly, for the number of shuffles, we generally choose around 10, as further increasing the shuffle count does not lead to notable gains.

MoM does not significantly impact efficiency since it only involves array computations. Ultimately, our total inference time is directly determined by the number of inference runs: it is simply the single-inference time multiplied by the number of runs. As shown in Table \ref{tab:ef}, even with dozens of inference runs, our approach remains more efficient than the single-inference times of most other baselines.

\section{Testing MAE for the multivariate setting}\label{sec:maes}
\begin{table*}[htbp]
\centering
\caption{
Testing MAE in the multivariate setting. The number in brackets indicates the rank. The best result is highlighted in bold, and the second best is underlined. CSDI runs out of memory on the Traffic and Electricity datasets.}
\resizebox{.98\textwidth}{!}{%
\begin{tabular}{c|lllllllll|l}%
\specialrule{1.1pt}{0pt}{0pt}
\ Method & \textit{\small NorPool} & \textit{\small Caiso} & \textit{\small Traffic} & \textit{\small Electricity} & \textit{\small Weather} & \textit{\small Exchange} & \textit{\small ETTh1} & \textit{\small ETTm1} & \textit{\small Wind} & Rank \\ 
\midrule
\textbf{Ours} & $\textbf{0.562}_{\raisebox{0.5ex}{\footnotesize (1)}}$ & $\underline{0.201}_{\raisebox{0.5ex}{\footnotesize (2)}}$ & $\textbf{0.254}_{\raisebox{0.5ex}{\footnotesize (1)}}$ & $\textbf{0.238}_{\raisebox{0.5ex}{\footnotesize (1)}}$ & $\underline{0.315}_{\raisebox{0.5ex}{\footnotesize (2)}}$ & $\textbf{0.079}_{\raisebox{0.5ex}{\footnotesize (1)}}$ & $\textbf{0.403}_{\raisebox{0.5ex}{\footnotesize (1)}}$ & $\textbf{0.363}_{\raisebox{0.5ex}{\footnotesize (1)}}$ & $\textbf{0.671}_{\raisebox{0.5ex}{\footnotesize (1)}}$ & $\textbf{1.22}$ \\

\midrule
mr-Diff & $0.604_{\raisebox{0.5ex}{\footnotesize (4)}}$ & $0.219_{\raisebox{0.5ex}{\footnotesize (5)}}$ & $0.320_{\raisebox{0.5ex}{\footnotesize (7)}}$ & $0.252_{\raisebox{0.5ex}{\footnotesize (5)}}$ & $0.324_{\raisebox{0.5ex}{\footnotesize (3)}}$ & $0.082_{\raisebox{0.5ex}{\footnotesize (5)}}$ & $0.422_{\raisebox{0.5ex}{\footnotesize (4)}}$ & $0.373_{\raisebox{0.5ex}{\footnotesize (3)}}$ & $\underline{0.675}_{\raisebox{0.5ex}{\footnotesize (2)}}$ & $4.22$ \\

TimeDiff & $0.611_{\raisebox{0.5ex}{\footnotesize (5)}}$ & $0.234_{\raisebox{0.5ex}{\footnotesize (8)}}$ & $0.384_{\raisebox{0.5ex}{\footnotesize (10)}}$ & $0.305_{\raisebox{0.5ex}{\footnotesize (8)}}$ & $\textbf{0.312}_{\raisebox{0.5ex}{\footnotesize (1)}}$ & $0.091_{\raisebox{0.5ex}{\footnotesize (9)}}$ & $0.430_{\raisebox{0.5ex}{\footnotesize (5)}}$ & $\underline{0.372}_{\raisebox{0.5ex}{\footnotesize (2)}}$ & $0.687_{\raisebox{0.5ex}{\footnotesize (4)}}$ & $5.78$ \\

TimeGrad & $0.821_{\raisebox{0.5ex}{\footnotesize (20)}}$ & $0.339_{\raisebox{0.5ex}{\footnotesize (19)}}$ & $0.849_{\raisebox{0.5ex}{\footnotesize (23)}}$ & $0.630_{\raisebox{0.5ex}{\footnotesize (22)}}$ & $0.381_{\raisebox{0.5ex}{\footnotesize (16)}}$ & $0.193_{\raisebox{0.5ex}{\footnotesize (20)}}$ & $0.719_{\raisebox{0.5ex}{\footnotesize (23)}}$ & $0.605_{\raisebox{0.5ex}{\footnotesize (23)}}$ & $0.793_{\raisebox{0.5ex}{\footnotesize (22)}}$ & $20.89$ \\

CSDI & $0.777_{\raisebox{0.5ex}{\footnotesize (18)}}$ & $0.345_{\raisebox{0.5ex}{\footnotesize (20)}}$ & - & - & $0.374_{\raisebox{0.5ex}{\footnotesize (14)}}$ & $0.194_{\raisebox{0.5ex}{\footnotesize (21)}}$ & $0.438_{\raisebox{0.5ex}{\footnotesize (7)}}$ & $0.442_{\raisebox{0.5ex}{\footnotesize (16)}}$ & $0.741_{\raisebox{0.5ex}{\footnotesize (11)}}$ & $15.29$ \\

SSSD & $0.753_{\raisebox{0.5ex}{\footnotesize (14)}}$ & $0.295_{\raisebox{0.5ex}{\footnotesize (11)}}$ & $0.398_{\raisebox{0.5ex}{\footnotesize (15)}}$ & $0.363_{\raisebox{0.5ex}{\footnotesize (13)}}$ & $0.350_{\raisebox{0.5ex}{\footnotesize (10)}}$ & $0.127_{\raisebox{0.5ex}{\footnotesize (14)}}$ & $0.561_{\raisebox{0.5ex}{\footnotesize (18)}}$ & $0.406_{\raisebox{0.5ex}{\footnotesize (11)}}$ & $0.778_{\raisebox{0.5ex}{\footnotesize (19)}}$ & $13.89$ \\

\midrule
D3VAE & $0.692_{\raisebox{0.5ex}{\footnotesize (11)}}$ & $0.331_{\raisebox{0.5ex}{\footnotesize (17)}}$ & $0.483_{\raisebox{0.5ex}{\footnotesize (18)}}$ & $0.372_{\raisebox{0.5ex}{\footnotesize (15)}}$ & $0.380_{\raisebox{0.5ex}{\footnotesize (15)}}$ & $0.301_{\raisebox{0.5ex}{\footnotesize (23)}}$ & $0.502_{\raisebox{0.5ex}{\footnotesize (15)}}$ & $0.391_{\raisebox{0.5ex}{\footnotesize (10)}}$ & $0.779_{\raisebox{0.5ex}{\footnotesize (20)}}$ & $16.00$ \\

CPF & $0.889_{\raisebox{0.5ex}{\footnotesize (22)}}$ & $0.424_{\raisebox{0.5ex}{\footnotesize (22)}}$ & $0.714_{\raisebox{0.5ex}{\footnotesize (22)}}$ & $0.643_{\raisebox{0.5ex}{\footnotesize (23)}}$ & $0.781_{\raisebox{0.5ex}{\footnotesize (24)}}$ & $0.082_{\raisebox{0.5ex}{\footnotesize (5)}}$ & $0.597_{\raisebox{0.5ex}{\footnotesize (21)}}$ & $0.472_{\raisebox{0.5ex}{\footnotesize (17)}}$ & $0.757_{\raisebox{0.5ex}{\footnotesize (16)}}$ & $19.11$ \\

PSA-GAN & $0.890_{\raisebox{0.5ex}{\footnotesize (23)}}$ & $0.477_{\raisebox{0.5ex}{\footnotesize (23)}}$ & $0.697_{\raisebox{0.5ex}{\footnotesize (21)}}$ & $0.533_{\raisebox{0.5ex}{\footnotesize (21)}}$ & $0.578_{\raisebox{0.5ex}{\footnotesize (23)}}$ & $0.087_{\raisebox{0.5ex}{\footnotesize (8)}}$ & $0.546_{\raisebox{0.5ex}{\footnotesize (17)}}$ & $0.488_{\raisebox{0.5ex}{\footnotesize (19)}}$ & $0.756_{\raisebox{0.5ex}{\footnotesize (14)}}$ & $18.78$ \\

\midrule
N-Hits & $0.643_{\raisebox{0.5ex}{\footnotesize (9)}}$ & $0.221_{\raisebox{0.5ex}{\footnotesize (6)}}$ & $0.268_{\raisebox{0.5ex}{\footnotesize (3)}}$ & $0.245_{\raisebox{0.5ex}{\footnotesize (4)}}$ & $0.335_{\raisebox{0.5ex}{\footnotesize (6)}}$ & $0.085_{\raisebox{0.5ex}{\footnotesize (7)}}$ & $0.480_{\raisebox{0.5ex}{\footnotesize (10)}}$ & $0.388_{\raisebox{0.5ex}{\footnotesize (8)}}$ & $0.734_{\raisebox{0.5ex}{\footnotesize (9)}}$ & $6.89$ \\

FiLM & $0.646_{\raisebox{0.5ex}{\footnotesize (10)}}$ & $0.278_{\raisebox{0.5ex}{\footnotesize (10)}}$ & $0.398_{\raisebox{0.5ex}{\footnotesize (15)}}$ & $0.320_{\raisebox{0.5ex}{\footnotesize (10)}}$ & $0.336_{\raisebox{0.5ex}{\footnotesize (7)}}$ & $\textbf{0.079}_{\raisebox{0.5ex}{\footnotesize (1)}}$ & $0.436_{\raisebox{0.5ex}{\footnotesize (6)}}$ & $0.374_{\raisebox{0.5ex}{\footnotesize (4)}}$ & $0.717_{\raisebox{0.5ex}{\footnotesize (6)}}$ & $7.67$ \\

Depts & $0.611_{\raisebox{0.5ex}{\footnotesize (5)}}$ & $0.204_{\raisebox{0.5ex}{\footnotesize (3)}}$ & $0.568_{\raisebox{0.5ex}{\footnotesize (20)}}$ & $0.401_{\raisebox{0.5ex}{\footnotesize (18)}}$ & $0.394_{\raisebox{0.5ex}{\footnotesize (18)}}$ & $0.100_{\raisebox{0.5ex}{\footnotesize (11)}}$ & $0.491_{\raisebox{0.5ex}{\footnotesize (13)}}$ & $0.412_{\raisebox{0.5ex}{\footnotesize (13)}}$ & $0.751_{\raisebox{0.5ex}{\footnotesize (13)}}$ & $12.67$ \\

NBeats & $0.832_{\raisebox{0.5ex}{\footnotesize (21)}}$ & $0.235_{\raisebox{0.5ex}{\footnotesize (9)}}$ & $\underline{0.265}_{\raisebox{0.5ex}{\footnotesize (2)}}$ & $0.370_{\raisebox{0.5ex}{\footnotesize (14)}}$ & $0.420_{\raisebox{0.5ex}{\footnotesize (19)}}$ & $0.081_{\raisebox{0.5ex}{\footnotesize (4)}}$ & $0.521_{\raisebox{0.5ex}{\footnotesize (16)}}$ & $0.409_{\raisebox{0.5ex}{\footnotesize (12)}}$ & $0.741_{\raisebox{0.5ex}{\footnotesize (11)}}$ & $12.00$ \\
\midrule
Scaleformer & $0.769_{\raisebox{0.5ex}{\footnotesize (17)}}$ & $0.310_{\raisebox{0.5ex}{\footnotesize (13)}}$ & $0.379_{\raisebox{0.5ex}{\footnotesize (9)}}$ & $0.304_{\raisebox{0.5ex}{\footnotesize (7)}}$ & $0.438_{\raisebox{0.5ex}{\footnotesize (20)}}$ & $0.138_{\raisebox{0.5ex}{\footnotesize (16)}}$ & $0.579_{\raisebox{0.5ex}{\footnotesize (20)}}$ & $0.475_{\raisebox{0.5ex}{\footnotesize (18)}}$ & $0.864_{\raisebox{0.5ex}{\footnotesize (23)}}$ & $15.89$ \\

PatchTST & $\underline{0.566}_{\raisebox{0.5ex}{\footnotesize (2)}}$ & $0.210_{\raisebox{0.5ex}{\footnotesize (4)}}$ & $0.276_{\raisebox{0.5ex}{\footnotesize (5)}}$ & $0.240_{\raisebox{0.5ex}{\footnotesize (3)}}$ & $0.326_{\raisebox{0.5ex}{\footnotesize (4)}}$ & $0.080_{\raisebox{0.5ex}{\footnotesize (3)}}$ & $\underline{0.416}_{\raisebox{0.5ex}{\footnotesize (2)}}$ & $0.378_{\raisebox{0.5ex}{\footnotesize (6)}}$ & $0.722_{\raisebox{0.5ex}{\footnotesize (7)}}$ & $\underline{4.00}$ \\

FedFormer & $0.744_{\raisebox{0.5ex}{\footnotesize (12)}}$ & $0.317_{\raisebox{0.5ex}{\footnotesize (14)}}$ & $0.385_{\raisebox{0.5ex}{\footnotesize (11)}}$ & $0.341_{\raisebox{0.5ex}{\footnotesize (12)}}$ & $0.347_{\raisebox{0.5ex}{\footnotesize (9)}}$ & $0.233_{\raisebox{0.5ex}{\footnotesize (22)}}$ & $0.484_{\raisebox{0.5ex}{\footnotesize (11)}}$ & $0.413_{\raisebox{0.5ex}{\footnotesize (14)}}$ & $0.762_{\raisebox{0.5ex}{\footnotesize (17)}}$ & $13.56$ \\

Autoformer & $0.751_{\raisebox{0.5ex}{\footnotesize (13)}}$ & $0.321_{\raisebox{0.5ex}{\footnotesize (15)}}$ & $0.392_{\raisebox{0.5ex}{\footnotesize (14)}}$ & $0.313_{\raisebox{0.5ex}{\footnotesize (9)}}$ & $0.354_{\raisebox{0.5ex}{\footnotesize (11)}}$ & $0.167_{\raisebox{0.5ex}{\footnotesize (17)}}$ & $0.484_{\raisebox{0.5ex}{\footnotesize (11)}}$ & $0.496_{\raisebox{0.5ex}{\footnotesize (20)}}$ & $0.756_{\raisebox{0.5ex}{\footnotesize (14)}}$ & $13.78$ \\

Pyraformer & $0.781_{\raisebox{0.5ex}{\footnotesize (19)}}$ & $0.371_{\raisebox{0.5ex}{\footnotesize (21)}}$ & $0.390_{\raisebox{0.5ex}{\footnotesize (12)}}$ & $0.379_{\raisebox{0.5ex}{\footnotesize (16)}}$ & $0.385_{\raisebox{0.5ex}{\footnotesize (17)}}$ & $0.112_{\raisebox{0.5ex}{\footnotesize (13)}}$ & $0.493_{\raisebox{0.5ex}{\footnotesize (14)}}$ & $0.435_{\raisebox{0.5ex}{\footnotesize (15)}}$ & $0.735_{\raisebox{0.5ex}{\footnotesize (10)}}$ & $15.22$ \\

Informer & $0.757_{\raisebox{0.5ex}{\footnotesize (15)}}$ & $0.336_{\raisebox{0.5ex}{\footnotesize (18)}}$ & $0.391_{\raisebox{0.5ex}{\footnotesize (13)}}$ & $0.383_{\raisebox{0.5ex}{\footnotesize (17)}}$ & $0.364_{\raisebox{0.5ex}{\footnotesize (12)}}$ & $0.192_{\raisebox{0.5ex}{\footnotesize (19)}}$ & $0.605_{\raisebox{0.5ex}{\footnotesize (22)}}$ & $0.542_{\raisebox{0.5ex}{\footnotesize (21)}}$ & $0.772_{\raisebox{0.5ex}{\footnotesize (18)}}$ & $17.22$ \\

Transformer & $0.765_{\raisebox{0.5ex}{\footnotesize (16)}}$ & $0.321_{\raisebox{0.5ex}{\footnotesize (15)}}$ & $0.410_{\raisebox{0.5ex}{\footnotesize (17)}}$ & $0.405_{\raisebox{0.5ex}{\footnotesize (19)}}$ & $0.370_{\raisebox{0.5ex}{\footnotesize (13)}}$ & $0.178_{\raisebox{0.5ex}{\footnotesize (18)}}$ & $0.567_{\raisebox{0.5ex}{\footnotesize (19)}}$ & $0.592_{\raisebox{0.5ex}{\footnotesize (22)}}$ & $0.785_{\raisebox{0.5ex}{\footnotesize (21)}}$ & $17.78$ \\

\midrule
SCINet & $0.601_{\raisebox{0.5ex}{\footnotesize (3)}}$ & $\textbf{0.193}_{\raisebox{0.5ex}{\footnotesize (1)}}$ & $0.335_{\raisebox{0.5ex}{\footnotesize (8)}}$ & $0.280_{\raisebox{0.5ex}{\footnotesize (6)}}$ & $0.344_{\raisebox{0.5ex}{\footnotesize (8)}}$ & $0.137_{\raisebox{0.5ex}{\footnotesize (15)}}$ & $0.463_{\raisebox{0.5ex}{\footnotesize (9)}}$ & $0.389_{\raisebox{0.5ex}{\footnotesize (9)}}$ & $0.732_{\raisebox{0.5ex}{\footnotesize (8)}}$ & $7.44$ \\

NLinear & $0.636_{\raisebox{0.5ex}{\footnotesize (7)}}$ & $0.223_{\raisebox{0.5ex}{\footnotesize (7)}}$ & $0.293_{\raisebox{0.5ex}{\footnotesize (6)}}$ & $\underline{0.239}_{\raisebox{0.5ex}{\footnotesize (2)}}$ & $0.328_{\raisebox{0.5ex}{\footnotesize (5)}}$ & $0.091_{\raisebox{0.5ex}{\footnotesize (9)}}$ & $0.418_{\raisebox{0.5ex}{\footnotesize (3)}}$ & $0.375_{\raisebox{0.5ex}{\footnotesize (5)}}$ & $0.706_{\raisebox{0.5ex}{\footnotesize (5)}}$ & $5.44$ \\

DLinear & $0.640_{\raisebox{0.5ex}{\footnotesize (8)}}$ & $0.497_{\raisebox{0.5ex}{\footnotesize (24)}}$ & $0.268_{\raisebox{0.5ex}{\footnotesize (3)}}$ & $0.336_{\raisebox{0.5ex}{\footnotesize (11)}}$ & $0.444_{\raisebox{0.5ex}{\footnotesize (21)}}$ & $0.102_{\raisebox{0.5ex}{\footnotesize (12)}}$ & $0.442_{\raisebox{0.5ex}{\footnotesize (8)}}$ & $0.378_{\raisebox{0.5ex}{\footnotesize (6)}}$ & $0.686_{\raisebox{0.5ex}{\footnotesize (3)}}$ & $10.67$ \\

LSTMa & $0.974_{\raisebox{0.5ex}{\footnotesize (24)}}$ & $0.305_{\raisebox{0.5ex}{\footnotesize (12)}}$ & $0.510_{\raisebox{0.5ex}{\footnotesize (19)}}$ & $0.444_{\raisebox{0.5ex}{\footnotesize (20)}}$ & $0.501_{\raisebox{0.5ex}{\footnotesize (22)}}$ & $0.534_{\raisebox{0.5ex}{\footnotesize (24)}}$ & $0.782_{\raisebox{0.5ex}{\footnotesize (24)}}$ & $0.699_{\raisebox{0.5ex}{\footnotesize (24)}}$ & $0.897_{\raisebox{0.5ex}{\footnotesize (24)}}$ & $21.44$ \\
\specialrule{1.1pt}{0pt}{0pt}
\end{tabular}
}
\label{tab:main_mae}
\end{table*}
Table \ref{tab:main_mae} shows the testing MAE results on the multivarate time series, where \textbf{SimDiff} also excels all competitive baselines.

\section{Limitations and Future Directions}\label{sec:lims}
Although SimDiff is lightweight and achieves extremely fast single-pass inference, it requires multiple (often dozens of) sampling runs to produce stable predictions. While this remains faster overall than prior diffusion models, reducing the number of required samples is a clear area for improvement. Future work could explore methods to achieve robust forecasts with only a few samples. Additionally, while SimDiff is designed for time series, extending its framework to other modalities remains a promising direction.


\end{document}